\documentclass{article}

\usepackage{microtype}
\usepackage{graphicx}
\usepackage{subfigure}
\usepackage{booktabs} 

\usepackage[hyphens]{url}
\usepackage{hyperref}
\hypersetup{breaklinks=true}

\usepackage[accepted]{icml2025}

\usepackage{amsmath}
\usepackage{amssymb}
\usepackage{mathtools}
\usepackage{amsthm}

\usepackage[capitalize,noabbrev]{cleveref}

\theoremstyle{plain}
\newtheorem{theorem}{Theorem}[section]
\newtheorem{proposition}[theorem]{Proposition}
\newtheorem{lemma}[theorem]{Lemma}

\theoremstyle{definition}

\newtheorem{assumption}[theorem]{Assumption}
\theoremstyle{remark}

\usepackage[textsize=tiny]{todonotes}

\usepackage{tikz}
\usetikzlibrary{arrows.meta} 
\usepackage{xcolor} 
\definecolor{darkgreen}{RGB}{40, 138, 65} 
\usepackage{caption}
\usepackage{subcaption} 
\usepackage{float}
\newcommand{\citetstar}[1]{\citeauthor{#1} \citetext{\citeyear{#1}*}}

\icmltitlerunning{Simple and Critical Iterative Denoising}

\usepackage{amsmath,amsfonts,bm}

\def\eqref#1{equation~\ref{#1}}

\def\1{\bm{1}}

\def\vone{{\bm{1}}}

\def\ve{{\bm{e}}}

\def\vh{{\bm{h}}}

\def\vm{{\bm{m}}}

\def\vx{{\bm{x}}}

\def\vz{{\bm{z}}}

\def\mA{{\bm{A}}}

\def\mE{{\bm{E}}}

\def\mI{{\bm{I}}}

\def\mQ{{\bm{Q}}}

\def\mW{{\bm{W}}}
\def\mX{{\bm{X}}}

\DeclareMathAlphabet{\mathsfit}{\encodingdefault}{\sfdefault}{m}{sl}
\SetMathAlphabet{\mathsfit}{bold}{\encodingdefault}{\sfdefault}{bx}{n}

\def\gE{{\mathcal{E}}}

\def\gG{{\mathcal{G}}}

\def\gL{{\mathcal{L}}}

\def\gU{{\mathcal{U}}}
\def\gV{{\mathcal{V}}}

\begin{document}

\twocolumn[
\icmltitle{Simple and Critical Iterative Denoising: \\ 
A Recasting of Discrete Diffusion in Graph Generation}



\icmlsetsymbol{equal}{*}

\begin{icmlauthorlist}
\icmlauthor{Yoann Boget}{heg,unige}
\end{icmlauthorlist}

\icmlaffiliation{heg}{Data Mining and Machine Learning group, School for Business Administration (HEG-Geneva), University of Applied Sciences and Arts of Western Switzerland}
\icmlaffiliation{unige}{Computer Science Department, University of Geneva, Switzerland}

\icmlcorrespondingauthor{Yoann Boget}{yoann.boget@hesge.ch}

\icmlkeywords{Machine Learning, ICML, Denoising, Discrete, Diffusion, Iterative, Critic, Critical, Graph, Generative, Model, Flow, Matching, Flow Matching, Compounding errors}

\vskip 0.3in
]



\printAffiliationsAndNotice{}  

\begin{abstract}
Discrete Diffusion and Flow Matching models have significantly advanced generative modeling for discrete structures, including graphs. However, the dependencies between intermediate noisy states lead to error accumulation and propagation during the reverse denoising process—a phenomenon known as \emph{compounding denoising errors}. To address this problem, we propose a novel framework called \emph{Simple Iterative Denoising}, which simplifies discrete diffusion and circumvents the issue by assuming conditional independence between intermediate states. 
Additionally, we enhance our model by incorporating a \emph{Critic}. During generation, the Critic selectively retains or corrupts elements in an instance based on their likelihood under the data distribution.
Our empirical evaluations demonstrate that the proposed method significantly outperforms existing discrete diffusion baselines in graph generation tasks.
\end{abstract}



\label{submission}

\section{Introduction}

Denoising models such as Discrete Diffusion and Discrete Flow Matching have significantly advanced generative modeling for discrete structures \citep{discdiff_austin, discdiff_campbell, Discrete_FlowMatching_campbell, discrete_flow_matching_gat}, including graphs \citep{discdiff_haefeli, digress}. Despite their success, these models suffer from a key limitation caused by the dependencies between intermediate noisy states in the noising and denoising processes. 
Errors introduced early in the process accumulate and propagate, degrading generative performance. 
This issue is particularly pronounced in mask diffusion, where the noising process progressively masks elements of an instance. 
The problem has been observed in discrete sequence modeling, where techniques such as \emph{corrector sampling} have been proposed to mitigate it.

\begin{figure}[H]
    \begin{center}
     \includegraphics[scale=0.25]{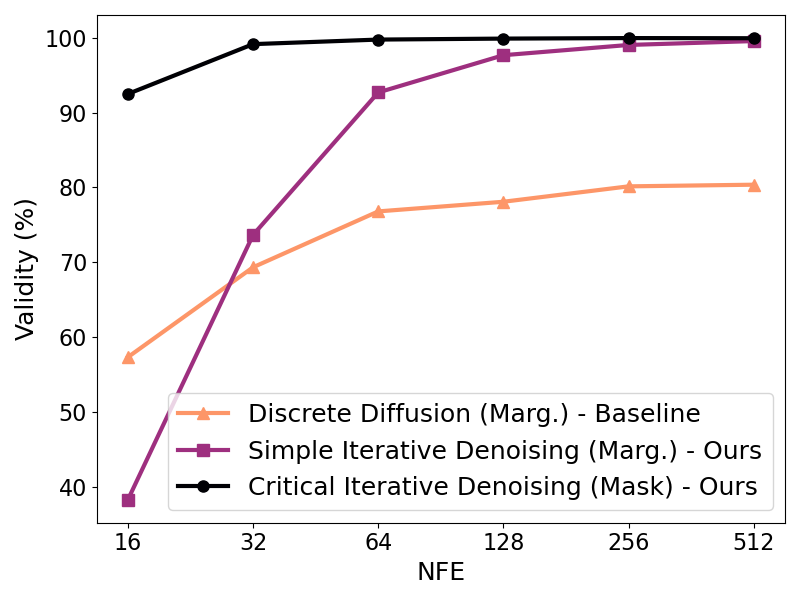}
    \end{center}
    \caption{Validity rate (without correction) of generated molecules trained on \texttt{Zinc250k} as a function of the Number of Function Evaluations (NFE) for three models: Discrete Diffusion (baseline), Simple Iterative Denoising (ours), and Critical Iterative Denoising (ours).}
    \label{fig:nfe_val_init}
\end{figure}

In graph modeling, marginal distributions over node and edge attributes have proven to be a more suitable noise distribution than the mask distribution, as corroborated by our experiments. However, we show both theoretically and empirically that \emph{compounding denoising errors} also affects discrete diffusion and flow matching models when the marginal distributions are used as the noise.

To address this challenge, we introduce a novel framework called \emph{Iterative Denoising}, which simplifies discrete diffusion by assuming that intermediate noisy states depend only on the clean data, making them conditionally independent of one another. By removing direct dependence on partially denoised instances from previous steps, our framework facilitates error correction and substantially improves generative performance.

Furthermore, we show that our model can be interpreted as selectively corrupting certain elements while leaving others untouched. Building on this insight, we introduce a \emph{Critic}, which modulates the corruption probability of each element based on the element's likelihood under the data distribution. 
We provide theoretical motivation for our approach and demonstrate empirically that the method further enhances models' performance.

Our empirical evaluations highlight the effectiveness of the proposed framework, demonstrating superior performance over existing discrete diffusion baselines in graph generation tasks. Notably, our method tackles the challenge of generating valid molecular structures. In Figure~\ref{fig:nfe_val_init}, we see that our method needs only a small number of denoising steps to reach almost 100\% molecule validity, while the standard Discrete Diffusion tops at 80\%. 

Our key contributions are as follows:

\begin{itemize}
    \item We identify \emph{compounding denoising errors} as a key limitation in discrete diffusion across various noise distributions, not just masking, as previously reported.

    \item To address this issue, we propose a novel, simple, and effective framework: \emph{Simple Iterative Denoising} (SID).

    \item We extend SID with a novel Critic mechanism that adjusts re-masking probabilities based on the predicted clean sample.

    \item Empirically, we show that our approach outperforms standard discrete denoising in graph generation tasks, with the Critic providing additional gains, especially in low-NFE settings.

\end{itemize}

Our code is available at: \href{https://github.com/yoboget/sid}{\texttt{github.com/yoboget/sid}}.
\section{Background}\label{sec:background}

Graph diffusion models are powerful methods for generating discrete graph structures, such as molecules. However, existing approaches, including Discrete Diffusion Models and Discrete Flow Matching (DFM), face challenges due to the temporal dependencies of the noising process. These dependencies cause error propagation and accumulation, ultimately degrading generative performance. This section introduces core concepts, discusses these limitations, and motivates our approach.

\subsection{Notation}

We define a graph as a set of nodes and edges, denoted by \(\gG = (\gV, \gE)\). A graph is represented by its adjacency matrix \(\mA \in [d_A+1]^{n \times n}\), where \(n = |\gV|\), and \([d_A + 1] = \{1, \dots, d_A + 1\}\), with \(d_A\) representing the number of edge types and the additional label corresponding to the absence of an edge. Node attributes, if present, are encoded as integers in an annotation vector \(\vx \in [d_X]^{n}\), where \(d_X\) is the number of node types.

To facilitate readability and simplify the transferability of the method to other discrete modalities, such as sequence modeling, we use \(z\) to denote a single element. In graph modeling, \(z\) may refer to either a node or an edge attribute when the formulation applies to both \(x^{(i)}\) and \(a^{(i, j)}\). The corresponding capital symbol \(Z\) represents the entire instance, such as a graph \(\gG\). For simplicity and consistency with common practice, we use \(p(x)\) to denote the Probability Mass Function (PMF) \(P(X=x)\). The distribution of a data point element is represented by the Dirac delta distribution, denoted as \(\delta_{z_1}(z)\), with all the mass concentrated on \(z_1\). 
We sometimes represent the univariate categorical distribution over the variable \(p(z)\) as a vector \(\vz\), where the $i$-th component $z_i$ denotes the probability that $z$ belongs to the category indexed by $i$.

We refer to \emph{mask distribution} as the distribution in which all the probability mass is concentrated on an additional synthetic attribute labeled as \texttt{Mask}, denoted by \(\delta_{\text{Mask}}(z)\). This distribution is sometimes referred to as the \emph{absorbing-state} distribution \citep{discdiff_austin}. We define \emph{mask diffusion} as the class of diffusion models that employ this distribution as noise.
We use \(q\) to refer to the noising distributions and \(p\) to the denoising ones.

\subsection{Discrete Diffusion Models}\label{sec:ddm}

In the forward process of diffusion models \citep{score-based_orginal, DDPM, scorebased_song_corrector_sampling}, each element transitions independently from the data distribution \(q_1(z) = \delta_{z_1}(z)\) at time \(t=1\) to a noise distribution \(q_0(z)\) at time \(t=0\), which contains (or tends to contain) no information about the original data.

The forward noising process is expressed under a Markovian assumption as:
\begin{align}
\label{eq:stdFwd}
q_{t|1}(z \mid z_1) = \prod_{r \in \tau} q_{r \mid r+\Delta_t}(z \mid z_{r+\Delta_t}),
\end{align}
where \(\tau = \{1-\Delta_t,\dots, t+\Delta_t, t\}\). 

The continuous-time case is described by taking \(\lim_{\Delta_t \to 0}q_{t|1}(z \mid z_1)\), which transforms the product into a geometric integral. 

For categorical (non-ordinal) discrete data, a popular family of noising distributions is given by:

\begin{equation}
    q_{t - \Delta_t \mid t}(z \mid z_t) = (1 - \beta_t)\, \delta_{z_t}(z) + \beta_t\, q_0(z).
\end{equation}
 
This formulation includes forward processes leading to a uniform noise (\(q_0(z) = \frac{1}{d_Z}  \;\forall z\)), mask noise (\( q_0(z) = \delta_{\text{Mask}}(z)\)), or a marginal noise (\( q_0(z) = m_z \), where \(m_z\) corresponds to the empirical proportion of elements with attribute $z$ in the dataset.

Under this formulation, the forward process admits a convenient closed-form solution to compute $q_{t|1}(z \mid z_1)$ efficiently:

\begin{equation}\label{eq:ddm_forward}
   q_{t|1}(z \mid z_1) = \alpha_t \delta_{z_1}(z) + (1-\alpha_t)q_0(z),
\end{equation}
where \(\alpha_t = \prod_{r \in \tau} (1-\beta_r)\). 

This noising process has been expressed using transition matrices in prior works, including \citet{discdiff_austin}, \citet{discdiff_campbell}, and \citet{digress} for graph-based formulations. 
We note that standard instantiations of DFM for categorical data with independent coupling \citep{discrete_flow_matching_gat} are equivalent to continuous-time discrete diffusion.
Therefore, the statements about discrete diffusion in the following also apply to these standard instantiations of DFM. 

By learning a backward process \(p^\theta_{s \mid t}(z \mid Z_t)\) for \(s = t + \Delta t\) and sampling iteratively from it, we generate data samples starting from the noise distribution.

\begin{figure}[H]
    \centering
    
    \subfigure[Discrete Diffusion.]
    {
        \begin{tikzpicture}[
            node distance=0.03\textwidth, 
            every node/.style={circle, draw, minimum size=0.8cm},
            >={Stealth[round]}, 
            shorten >=1pt 
        ]
        
        \node (z1) at (0, 0) {$z_{1}$};
                \node[below=0.01cm of z1, draw=none, rectangle, font=\small] (label) {Data};
        \node (dots) [right=of z1] {$...$};
        \node (zt1) [right=of dots] {$z_{s}$}; 
        \node (zt) [right=of zt1] {$z_t$}; 
        \node (dots2) [right=of zt] {$...$};
        \node (z0) [right=of dots2] {$z_0$}; 
         \node[below=0.01cm of z0, draw=none, rectangle, font=\small] (label) {Noise};

        \draw[->, blue] (z1) -- (dots);
        \draw[->, blue] (dots) -- (zt1);
        \draw[->, blue] (zt1) -- (zt); 
        \draw[->, blue] (zt) -- (dots2);
        
        \end{tikzpicture}
        \label{fig:noising_a}
        }

\subfigure[Iterative Denoising.]
{   
        \begin{tikzpicture}[
            node distance=0.03\textwidth, 
            every node/.style={circle, draw, minimum size=0.8cm, font=\large},
            >={Stealth[round]}, 
            shorten >=1pt 
        ]
        \node (z1) at (0, 0) {$z_{1}$};
        \node[below=0.01cm of z1, draw=none, rectangle, font=\small] (label) {Data};

        \node (dots) [right=of z1] {$...$};
        \node (zt1) [right=of dots] {$z_{s}$}; 
        \node (zt) [right=of zt1] {$z_t$}; 
        \node (dots2) [right=of zt] {$...$};
        \node (z0) [right=of dots2] {$z_0$}; 
        \node[below=0.01cm of z0, draw=none, rectangle, font=\small] (label) {Noise};

        \draw[->, blue] (z1) -- (dots);
        \draw[->, blue] (z1) to[bend left=30]  (zt1);
        \draw[->, blue] (z1) to[bend left=30]  (zt); 
        \draw[->, blue] (z1) to[bend left=30]  (dots2);

        \end{tikzpicture}  
        \label{fig:noising_b}
    }
    \caption{Noising Graphical Models in \subref{fig:noising_a} Discrete Diffusion: $z_t \sim \color{blue}q_{t \mid s}(z \mid z_{s})$; and \subref{fig:noising_b} Iterative Denoising: $z_t \sim \color{blue}q_{t \mid 1}(z \mid z_{1})$}
    \label{fig:noising}
    
\end{figure}

\subsection{Denoising Dependencies and Compounding Error}\label{sec:time_dep_issue}

At the beginning of the discrete diffusion backward process, \(Z_t\) contains limited information, resulting in a high-entropy distribution \(p^\theta_{s \mid t}(z \mid Z_t)\), from which the elements \(z_s\) of \(Z_s\) are independently sampled. At time \(s\), the joint distribution from the backward process:
\[
p^\theta_{s \mid t}(Z \mid Z_t) = \prod_{z \in Z} p^\theta_{s \mid t}(z \mid Z_t)
\]
may deviate significantly from the forward distribution due to early-stage denoising errors. 
This leads to error accumulation, a phenomenon known as the \emph{compounding denoising error} \citep{compounding_denoising_error}.

The compounding denoising error is particularly evident in mask diffusion, where unmasked elements cannot be masked again, as the conditional distribution collapses to a Dirac delta:
\begin{equation}
    p^\theta_{s \mid t}(z \mid Z_t) = \delta_{z_t}(z), \quad \forall s \geq t \quad \text{if} \quad z_t \neq \text{Mask}.
\end{equation}

While the issue is particularly pronounced in mask diffusion, we argue that it also affects other noising distributions $q_0$ such as the uniform and marginal distributions. 

In these cases, the compounding denoising error is somewhat less severe as all elements can change at each denoising step, allowing for corrections. However, \emph{due to the direct dependence of $z_s$ on $z_t$} (red arrow in Figure~\ref{fig:denoising}~\subref{fig:denoising_a}) the probability of correction is low, and the risk of error propagation and accumulation remains significant.
Specifically, \citet{discrete_flow_matching_gat} show that a denoising step in discrete diffusion can be expressed as:
\begin{multline}
    p_{t+\Delta_t|t}(z \mid Z_t) = \delta_{z_t}(z) + \\ \Delta_t\frac{\dot{\alpha_t}}{1 - \alpha_t} \left[ p_{1|t}(z \mid Z_t) - \delta_{z_t}(z) \right], 
\end{multline}
where $\dot{\alpha_t}$ is the $\alpha_t$ time-derivative.  
The second term on the right-hand side represents the probability of modifying an element in a denoising step of size \(\Delta_t\). The scaling factor \(\Delta_t \frac{\dot{\alpha_t}}{1 - \alpha_t}\), in particular, imposes a strong constraint on the probability of modifying an element. Consequently, \emph{compounding denoising errors} are not limited to mask diffusion, but also affects other common noise distributions, such as uniform and marginal distributions.  

\subsection{Motivation}

We argue that direct dependence on previous intermediate states in existing models hinders the generative performance of discrete diffusion models for graph generation. These models still struggle to consistently generate graphs with desired structural properties, such as valid molecular graphs or planar graphs.

In this work, we propose a simple yet effective framework that mitigates compounding denoising errors by removing dependencies on previous intermediate states in the noising process. Unlike standard DDM and DFM, which rely on the Markovian assumption, our Iterative Denoising approach (Sec. \ref{sec:ID}) assumes conditional independence from prior noisy states, alleviating the problem of error accumulation. While our method operates in a discrete setting, it relates to the general degrade-and-restore strategy developed for image generation\citep{cold_diff}. 

Additionally, we introduce a Critic-Guided Sampling procedure (Sec. \ref{sec:CID}), which prioritizes re-noising elements with lower probabilities under the data distribution, further improving generative performance, especially in molecular graph generation.

\subsection{Related works}

Our work primarily relates to two areas of research: methods that address compounding denoising errors and generative models for graphs. In the following, we review relevant contributions in both fields.

\subsubsection{Addressing Compounding Denoising Errors}

Discrete diffusion approaches to mitigating compounding denoising errors follow two main strategies: corrector sampling and planning.
Corrector sampling (also known as predictor-corrector sampling) \citep{Discrete_FlowMatching_campbell, compounding_denoising_error} involves alternating denoising (backward) steps with noising (forward) steps to improve convergence. Recent work \citep{discrete_flow_matching_gat} integrates denoising and corrector sampling by simultaneously performing $k$ backward and $k-1$ forward steps. In the context of mask diffusion, this method allows for the correction of previously sampled elements. In Section \ref{sec:ID}, we show that our \emph{Simple Iterative Denoising} process is formally equivalent to corrector sampling with maximal corrector steps. However, because elements are randomly re-noised, the efficiency of this approach remains limited. Extending this framework, our Critical Iterative Denoising (CID) introduces modulated re-masking probabilities, effectively acting as a probabilistic planner for re-masking decisions. Critically, unlike prior work, CID operates directly on fully denoised predictions.

Planning, as defined by \citet{think_generate}, involves selecting which elements to denoise (i.e., unmask) at each iteration. This selection can be guided by confidence scores from model outputs \citetext{\citealp{informed_correctors}, \citealp{repara_disc_diff_zheng}, \citealp{LLDM}*} or by a dedicated planner \citetext{\citealp{think_generate}; \citealp{plan_for_the_best}*}\footnote{References marked with '*' denote concurrent work, meaning they were published after this article was submitted.}. Outside discrete diffusion, confidence-based unmasking has been explored by \citet{maskgit}. Notably, planning does not influence the model's capacity to revise previously sampled elements. Using mask diffusion, \citetstar{planning_for_mask_diffusion} mitigate this issue by jointly deciding which masked elements to unmask and which unmasked elements to remask.
In contrast, our Simple Iterative Denoising approach removes the need to explicitly plan a denoising order by predicting all elements simultaneously at each step. 
Furthermore, our approach generalizes beyond masking to support various noising distributions.

\subsubsection{Graph Generative Models}

A key advantage of discrete denoising methods, including discrete diffusion, discrete flow matching, and our iterative denoising, is that they do not depend on a specific element ordering. This property offers a flexible framework for sequence infilling and generation of unordered structures such as sets and graphs. In this work, we focus on the application of denoising models for graph generation.

Recent denoising generative models for graphs leverage the order-agnostic nature of denoising processes to maintain permutation equivariance. Some approaches assume continuous diffusion spaces—such as score-based models \citep{edp-gnn, gdss}, diffusion bridges \citep{drum}, and flow matching \citep{VarFlowMatching}—but this assumption breaks the discrete nature of graph structures during noising. Discrete diffusion models, whether in discrete time \citep{discdiff_haefeli, digress} or continuous time \citep{continuous_time_graph_diffusion}, and discrete flow matching models, have shown strong performance on graph generation. However, these models are affected by compounding denoising errors. Our approach mitigates this issue, leading to improved generative performance. A broader review of graph generative models, including non-equivariant approaches, is provided in Appendix \ref{ap:related_works}.

\begin{figure}[H]
    \centering
    \subfigure[Discrete Diffusion]
    {
        \begin{tikzpicture}[
            node distance=0.03\textwidth, 
            every node/.style={circle, draw, minimum size=0.8cm, font=\large},
            >={Stealth[round]}, 
            shorten >=1pt 
        ]
        
        \node (z1) at (0, 0) {$z_{1}$};
        \node (dots) [right=of z1] {$...$};
        \node (zt1) [right=of dots] {$z_{s}$}; 
        \node (zt) [right=of zt1] {$Z_t$}; 
        
        \draw[->, darkgreen] (zt) to[bend left=30]  (z1); 
        \draw[->, blue] (z1) to[bend left=30]  (zt1);
        \draw[->, red] (zt) --  (zt1);
        \end{tikzpicture} 
        \label{fig:denoising_a}
    }
    \vspace{0.5cm}
    \subfigure[{Simple Iterative Denoising}]
    {
        \begin{tikzpicture}[
            node distance=0.03\textwidth, 
            every node/.style={circle, draw, minimum size=0.8cm, font=\large},
            >={Stealth[round]}, 
            shorten >=1pt 
        ]
        
        \node (z1) at (0, 0) {$z_{1}$};
        \node (dots) [right=of z1] {$...$};
        \node (zt1) [right=of dots] {$z_{s}$}; 
        \node (zt) [right=of zt1] {$Z_t$}; 
        
        \draw[->, darkgreen] (zt) to[bend left=30]  (z1); 
        \draw[->, blue] (z1) to[bend left=30]  (zt1);
        \end{tikzpicture}
        \label{fig:denoising_b}
        }
    \caption{Denoising Graphical Model for a single denoising step in \subref{fig:denoising_a} Discrete Diffusion:  
    $\hat{z}_s \sim p_{s \mid t}(z \mid Z_t) = \sum_{z_1} p_{s|1, t}(z \mid \color{blue}z_1, \color{red}z_t\color{black})\color{darkgreen}p_{1|t}(z \mid Z_{t})$; and in \subref{fig:denoising_b} Simple Iterative Denoising  $\hat{z}_s \sim p_{s \mid t}(z \mid Z_t)=\sum_{z_1} \color{blue}q_{s|1}(z \mid z_1)\color{darkgreen}p_{1|t}(z \mid Z_{t})$.
    }
    
    \label{fig:denoising}
\end{figure}

\section{Simple Iterative Denoising}\label{sec:ID}

In this section, we introduce Simple Iterative Denoising (SID), a simple yet effective generative framework for modeling discrete-state graphs. 
This approach simplifies existing denoising models, such as DDM and DFM, while enhancing generative performance. 
SID serves as the foundation for Critical Iterative Denoising (CID), which further improves the generative performance.

Similar to DDM, SID consists of both noising and denoising operations. The generative sampling procedure involves iteratively applying small denoising steps, enabling effective graph generation.

Notably, if we have access to a pre-trained DDM or DFM denoiser, i.e., a parameterized model predicting $p^\theta_{1|t}(z \mid Z_t) $, the SID method is training-free.
Table \ref{tab:compar_summary} summarizes the key differences between SID, DDM, and DFM.
In the following, we present the method in detail. 

\subsection{Noising}

We define a noising process that independently acts on each element of an instance as follows:
\begin{equation}\label{eq:noising}
 q_{t|1}(z \mid z_1) = \alpha_t \delta_{z_1}(z) + (1-\alpha_t)q_0(z), 
\end{equation}
where \( q_0(z) \) represents the noise distribution \( \delta_{z_1}(z) \) is the data distribution, and \( \alpha_t \) is a non-decreasing scheduling parameter in \([0, 1]\). 
The scheduler \(\alpha_t\), which defines the noise level, depends on a continuous time parameter, which also takes values in \([0, 1]\). 

Unlike standard diffusion models, which compute noise distributions via a Markov process (Eq. \ref{eq:stdFwd}) such that \( q_{t|t+\Delta_t, 1}(z \mid z_{t+\Delta_t}, z_1) = q_{t|t+\Delta_t}(z \mid z_{t+\Delta_t})\), our intermediate noisy distributions are directly parameterized as a mixture of the noise and the data distribution.
Thus, we do not assume any dependencies in the noising process other than the dependency on the original data point $z_1$ and on the noise level. 
This assumption introduces a significant simplification over DDM and DFM, leading to beneficial implications for the SID denoising process. In Figure~\ref{fig:noising}, we juxtapose the graphical models of our method and DDM at the noising phase. More formally we assume conditional independence of the noisy distributions across time given $z_1$:
\begin{assumption}\label{as:conditional_indep} 
The intermediate noisy distribution at time $t$ is conditionally independent from other intermediate states such that: 
 \begin{equation}\label{eq:conditional_indep}
 q_{t|1}(z \mid z_1)= q_{t|1}(z \mid z_s, z_1) \quad \forall \: t\neq s.
\end{equation} 
\end{assumption}
As a result, our noising procedure \emph{is not} a diffusion process in the traditional sense, as the state at time $t$ depends only on the initial state and not on the noising trajectory or previous intermediate states. 
Under this assumption, the denoising phase does not suffer from error propagation and accumulation caused by the direct dependencies to previous states inherent in standard DDM and DFM (see Section \ref{sec:denoising}).

Despite this assumption, we note that the distribution at time $t$ in our method is identical to that of discrete diffusion, as formalized in the following proposition:
\begin{proposition}\label{prop:forward} 
 Given a noise distribution $q_0$ and a scheduler $\alpha(t)$, the distribution $q^{SID}_{t|1}(z \mid z_1)$ defined by our noising procedure (Equation \ref{eq:noising}) is equal to the distribution $q^{DDM}_{t|1}(z \mid z_1)$ obtained by the discrete forward process distribution defined in Equation \ref{eq:stdFwd}. 
\end{proposition}

\begin{proof}
 See Appendix \ref{ap:proof_forward}
\end{proof}
The proposition follows from the fact that our noising procedure corresponds to the closed-form expression (Eq. \ref{eq:ddm_forward}) used in discrete diffusion to sample efficiently from $q_{t|1}(z \mid z_1)$. 
However, we make no assumption that the noisy distribution results from a diffusion process. 

\subsection{Denoising}\label{sec:denoising}

Denoising progressively refines noisy instances by learning the distribution \(p_{s|t}(z \mid Z_t)\), where $s = t + \Delta t$. 
By leveraging Assumption \ref{as:conditional_indep}, the denoising process is significantly simplified compared to discrete diffusion, as formalized in the following proposition:
\begin{proposition}
 Given the noising process defined in Equation \ref{eq:noising} and Assumption \ref{as:conditional_indep}, the denoising process is expressed as: 
 \begin{equation}\label{eq:denoising}
 p_{s|t}(z \mid Z_{t}) = \alpha_s p_{1|t}(z \mid Z_t) + (1-\alpha_s) q_0(z),
 \end{equation} 
\end{proposition}
\begin{proof}
See Appendix \ref{ap:proofs_denoising}.
\end{proof}
Under Assumption \ref{as:conditional_indep}, the denoising process factorizes as:
 \begin{equation}\label{eq:denoising_fact}
 p_{s|t}(z \mid Z_{t}) = \sum_{z_1} p_{1|t}(z_1 \mid Z_{t})q_{s|1}(z \mid z_1).
 \end{equation} 
Thus, we interpret the denoising as a two-step process: 1. Predict a clean instance from the noisy data, \(p_{1|t}(z_1 \mid Z_{t})\), 2. Re-noise the predicted clean instance, \(q_{s|1}(z \mid z_1)\). The key difference from the standard DDM model is that $z_s$ in SID directly depends only on the predicted clean instance, but not on $z_t$. In Figure \ref{fig:denoising} we juxtapose the denoising graphical models of DDM and SID. The fact that $z_s$ is generated only from the predicted clean instance acts as
a barrier to error propagation. 
This factorization will also prove useful in Section \ref{sec:CID}.

At inference time, we define the Number of Function Evaluations (NFE), denoted as $T$, as a hyperparameter. The time step size is set to $\Delta_t = 1/T$. The whole denoising process consists of iterating $T$ times the denoising step defined in Equation \ref{eq:denoising}.

\subsubsection{Corrector Sampling Interpretation}
The denoising process can also be viewed through the lens of discrete diffusion corrector sampling. Defining a discrete diffusion corrector sampling step as applying $k\Delta_t$ backward steps and $(k-1)\Delta_t$ forward steps, and noticing that the maximal size for corrector step at time $t$ is $k\Delta_t=1-t$, we state the following proposition:
\begin{proposition}\label{prop:corrector}
 A \emph{Simple Iterative Denoising} step, as described in Equation \ref{eq:denoising}, is equivalent to a discrete diffusion corrector sampling with maximal corrector step sizes.
\end{proposition}
\begin{proof} 
See Appendix \ref{ap:proofs_corrector}. 
\end{proof}
Consequently, the denoising process inherits the properties of the backward process in discrete diffusion; by iteratively sampling $p_{s|t}(z \mid Z_{t})$ with sufficiently small $\Delta_t$, the denoising process gradually transitions from the noise distribution to the data distribution. 
Our mask SID also inherits the time-independence property of mask denoisers shown by \citet{time_indep_absorbing} and \citet{simple_and_effective_MDL}.

\subsection{Parametrization and Learning}

As in discrete diffusion, the conditional probability $p_{1|t}(z_1 \mid Z_t)$ is intractable. We model this distribution by parametrizing a neural network $f_\theta(Z_t, \alpha_t)$ referred to as the \emph{denoiser}.
We use Graph Neural Networks (GNNs) to enforce equivariance to node permutations, thereby preserving the structural invariance of graphs.

Since the training objective matches that of DDM and DFM, we can adopt any of the training criteria proposed for these models to train $f_\theta(Z_t, \alpha_t)$. Following Digress \citep{digress}, we minimize the weighted negative log-likelihood in our experiments: 

\begin{align}
\label{eq:loss}
 \gL =& \mathbb{E}_{\gG \sim p_{\text{data}}, t\sim\gU(0, 1), \gG_{t} \sim (q_{t|1}(x_t^{(i)} \mid x^{(i)}_1), 
 q_{t|1}(e^{(i, j)}_t \mid e^{(i, j)}_1))} \\ 
 & \biggl[ \biggl. \gamma 
 \sum_{x^{(i)}_1}
 \left[-\log(p_\theta(x^{(i)}_1 \mid \gG_{t})) \right] \nonumber \\ 
 & 
 + (1-\gamma) 
 \sum_{e^{(i, j)}_1}
 \left[-\log(p_\theta(e^{(i, j)}_1| \gG_{t}))\right] \biggr. \biggr] \nonumber, 
\end{align}
where $\gamma$ is a weighting factor between nodes and edges. We use $\gamma = n/(n+m)$. At inference time, Simple Iterative Denoising can leverage any pre-trained DDM model.

\section{Critical Iterative Denoising}\label{sec:CID}

This section introduces \emph{Critical Iterative Denoising} (CID), a method designed to improve \emph{Simple Iterative Denoising}.

As discussed in Section \ref{sec:ID}, Mask SID can be viewed as a sequence of denoising steps comprising: (1) unmasking elements of an instance based on the denoiser's prediction and (2) re-masking a random subset of the instance's elements. 

However, in Mask SID, all elements are re-masked with the same probability. This approach is suboptimal, as not all elements are equally likely under the data distribution after the unmasking step.

CID addresses this limitation by dynamically adjusting the re-noising probability during sampling. 
Specifically, CID increases the re-noising probability for predicted denoised elements overrepresented under the data distribution and decreases it for those underrepresented.
CID serves two goals:
(1) Reducing error propagation by resampling elements that are likely out of the data distribution, and 
(2) Accelerating inference by decreasing the number of function evaluations required during sampling. 
As shown in Figure \ref{fig:nfe_val_init}, CID achieves a state-of-the-art validity rate on \texttt{Zinc250k} with only a few dozen of denoising steps.

\subsection{Preliminaries}\label{sec:cid_prelim}

Before describing the method, we recast the definition of the noise
distributions by introducing a Bernoulli random variable $a_t$, whose mean is given by the scheduling parameter $\alpha_t$ from Equation \ref{eq:noising}; $a_t$ indicates whether an element has been corrupted ($a_t=0$) or not ($a_t=1)$. 
As with \(z\), we use the capital letter \(A\) to denote the set of indicators for all elements in an instance. 
This formulation allows us to interpret the noising process as selectively
corrupting some elements while leaving others untouched. 
Through the random variables $a_t$, we can identify
which elements $z_t$ have been noised.
This identification would otherwise be impossible for any noise distribution sharing support with the data distribution (i.e., all practical noise distributions except the mask distribution). 

Given the random variable $a_t \sim p_{\alpha_t}(a)=\text{Bernoulli}(a; \alpha_t)$, we rewrite Equation \ref{eq:noising} as:
\begin{equation}\label{eq:fact_noising}
 q_{t|1}(z|z_1) = p_{\alpha_t}(a) \delta_{z_1}(z) + 
 (1-p_{\alpha_t}(a)) q_0(z).
\end{equation}

Similarly, we write the denoising equation as:
\begin{equation} 
\label{eq:fact_denoising}
 p_{1|t}(z \mid Z_t, A_t) = a_t \delta_{z_1}(z) + (1-a_t) p^\theta_{1|t}(z \mid Z_{t},A_t)
\end{equation} 
Intuitively, when the element is not corrupted ($a_t=1$), it remains unchanged ($\delta_{z_1}(z)$); when $a_t=0$ we predict it using the denoiser. 
For simplicity, we define $p_{\text{data}}(z) := \delta_{z_1}(z)$ and 
\(p_{\text{pred}}(z):= p^\theta_{1|t}(z \mid Z_{t},A_t)\).

In principle, our method supports any noise distribution and produces pairs $(z_t, a_t)$, which indicate whether the element $z_t$ was sampled from the noise distribution or corresponds to the original clean element. However, denoising using the pair $(z_t, a_t)$ reduces denoising models with any noise distribution (e.g., marginal, uniform) to a mask denoising model.

To see this, note that the corruption indicator \(a_t\) acts as a mask, with \(a_t = 0\) indicating a masked element. Defining the set of uncorrupted elements as \(Z^a_t = \{z_t \in Z_t \mid a_t = 1\}\), and its complement as \(Z^{\bar{a}}_t\), the masked denoiser can be expressed as \(p^\theta_{1 \mid t}(z \mid Z^a_t, A_t)\). Since \(Z^{\bar{a}}_t\) carries no information about \(Z_1\), it follows that: 
$
p^\theta_{1 \mid t}(z \mid Z_t, A_t) = p^\theta_{1 \mid t}(z \mid Z^a_t, A_t).
$
Because Critical Iterative Denoising (CID) explicitly requires \(A_t\), and any noise distribution under this configuration collapses to the Mask Simple Iterative Denoising (Mask SID) model, we always use Mask SID within CID.

\subsection{Iterative Denoising with a Critic}

We now describe the \emph{Critic}, which predicts the corruption state of graph elements during denoising.
Let $\hat{z}_{1|t} \sim p_{1|t}(z \mid Z_{t}, A_t)$ denote a denoised element at time $t$.
CID trains a Critic $C$ to estimate $\hat{\alpha}_t = p_\phi(a \mid \hat{Z}_{1|t})$, the probability that \(\hat{z}\) originates from $p_{\text{data}}$ rather than $p_{\text{pred}}$. The Critic is trained by minimizing the negative log-likelihood of the corruption indicators $a^{(i)}_t$ produced during the noising phase (one per element and time step):
\begin{equation}\label{eq:critic_objective} 
 \gL_\phi = -\mathbb{E}_{t, \hat{Z}_{1|t}} \sum_i \log p_\phi(a^{(i)}_t \mid \hat{Z}_{1|t}), 
\end{equation}
During inference, $\hat{\alpha}_t$ parametrizes a Bernoulli distribution $p_{\hat{\alpha}_t}(a_t) = \text{Bernoulli}(a; \hat{\alpha}_t)$ which determines the noising probability 
$1-\hat{\alpha}_t$. Recall that $\alpha_t$ is the scheduler's noise rate at time step $t$. Thus, the Critic's output can be interpreted as an adaptive, element-wise noise schedule.

The following theorem characterizes the optimal critic:
\begin{theorem}\label{prop:optimal_critic}
 The optimal Critic $C^*$ is:
 \begin{equation}\label{eq:optimal_critic}
 C^*(\hat{z}_{1|t}) = 
 \frac{\alpha_t p_{\text{data}}(\hat{z}_{1|t})}{\alpha_t p_{\text{data}}(\hat{z}_{1|t}) + (1-\alpha_t)p_{\text{pred}}(\hat{z}_{1|t})}
 \end{equation}
\end{theorem}
\begin{proof}
 See Appendix \ref{ap:proofs}
\end{proof}
From Theorem \ref{prop:optimal_critic}, two lemmas follow:
\begin{lemma}
 If $p_{\text{data}}(\hat{z}_{1|t}) = p_{\text{pred}}(\hat{z}_{1|t})$, the optimal $\hat{\alpha}^*_t$ coincide with the \emph{true} $\alpha_t$, that is:
 \begin{equation}
 p_{\text{data}}(\hat{z}_{1|t}) = p_{\text{pred}}(\hat{z}_{1|t}) \implies \hat{\alpha}^*_t =\alpha_t
 \end{equation}
\end{lemma}
\begin{proof}
 The lemma follows directly from Theorem \ref{prop:optimal_critic}.
\end{proof}
\begin{lemma}\label{prop:alphas_compar}
 If $p_{\text{data}}(\hat{z}_{1|t}) > p_{\text{pred}}(\hat{z}_{1|t})$, the optimal Critic's noising probability 
 $\hat{\alpha}_t^*$ is smaller than the schedule's noising rate $\alpha_t$, that is:
 \begin{equation}
 p_{\text{data}}(\hat{z}_{1|t}) > p_{\text{pred}}(\hat{z}_{1|t}) \implies \alpha_t > \hat{\alpha}_t^*. 
 \end{equation}
 Conversely, if $p_{\text{data}}(\hat{z}_{1|t}) < p_{\text{pred}}(\hat{z}_{1|t})$, the optimal Critic's noising probability 
 $\hat{\alpha}_t^*$ is larger than the schedule's noising rate $\alpha_t$:
 \begin{equation}
 p_{\text{data}}(\hat{z}_{1|t}) < p_{\text{pred}}(\hat{z}_{1|t}) \implies \alpha_t < \hat{\alpha}_t^*.
 \end{equation}
\end{lemma}
\begin{proof}
 See Appendix \ref{ap:alpha_compar}
\end{proof}
Additionally, if \(p_{\text{pred}}(\hat{z}_{1|t}) = 0 \), then $\hat{z}_{1|t}$ will not be masked (\(\alpha_t = 0\)). Conversely, if $p_{\text{data}}(\hat{z}_{1|t}) = 0$, the element is out of distribution and will be masked with probability \(\alpha_t = 1\).

Thanks to Theorem \ref{prop:optimal_critic} and the following lemmas, the Critic steers the denoised element toward the data distribution, by re-noising with higher probability those elements that are overrepresented under the data distribution, while preserving underrepresented ones.

\subsection{Implementation and Sampling}
Since $p_{\alpha_t}(a)$ (Equation \ref{eq:fact_noising}) does not depend on $\hat{Z}_{1|t}$, $\mathbb{E}_{\hat{Z}_{1|t}} p(a \mid \hat{Z}_{1|t}) = p_{\alpha_t}(a) = \alpha_t$, 
we actually cast the Critic as a predictor of the residual logit with respect to the true $\alpha_t$:
\begin{equation}
 p_\phi(a \mid \hat{Z}_{1|t}) = \sigma(f_\phi(\hat{Z}_{1|t}, \alpha_t) + \sigma^{-1}(\alpha_t)), 
\end{equation}
where $\sigma$ is the sigmoid function and $f_\phi(\hat{Z}_{1|t}, \alpha_t)$ is a GNN. 
The Critic operates on the outputs of a fixed denoiser and is trained \emph{post hoc}, requiring no retraining of the denoiser itself. 
We can leverage any denoiser implementing mask diffusion and no retraining is needed. 

During inference, we are interested in $\hat{\alpha}_{t+\Delta_t}$ rather than $\hat{\alpha}_{t}$. We therefore use the approximation $\hat{\alpha}_{t + \Delta_t} \approx \sigma(f_\phi(\hat{Z}_{1|t}, \alpha_t) + \sigma^{-1}(\alpha_{t + \Delta_t}))$.

In summary, each denoising step involves: 1. sampling $\hat{Z}_{1|t}$ from the Denoiser, 2. computing $\hat{\alpha}_{t + \Delta_t}$ via the Critic, 3. re-noising using Equation $\ref{eq:noising}$, with $\alpha_t$ replaced by $\hat{\alpha}_{t + \Delta_t}$. 

\section{Evaluation}\label{sec:eval}

We evaluate our model on both molecular and synthetic graph datasets. 

For molecular data, we use the \texttt{QM9} and \texttt{ZINC250k} datasets. 
The \texttt{QM9} dataset contains 133,885 molecules with up to 9 atoms of 4 types, while the \texttt{ZINC250k} dataset consists of 250,000 molecular graphs with up to 38 heavy atoms of 9 types.
For generic graphs, we run experiments on the \texttt{Planar} and \texttt{Stochastic Block Model} (SBM) datasets. Both contain 200 unattributed graphs with 64 nodes and up to 200 nodes, respectively.
Visualizations of generated molecules and graphs are available in Appendix \ref{ap:visual}.

Our objective is to evaluate the proposed method through direct comparison with Discrete Diffusion Models (DDMs). To ensure a rigorous comparison, we train two denoisers, one using the marginal distribution and the other using the mask distribution. We use these denoisers to compare the two sampling procedures: Discrete Diffusion and Simple Iterative Denoising. Additionally, we train a Critic for the model using the mask distribution.
Specifically, the Marginal DDM and Marginal SID share the same denoiser, while the Mask DDM, Mask SID, and Mask CID use a another shared denoiser with identical architecture and parameters.

This setup ensures a controlled and fair ablation study, isolating the impact of our iterative denoising approach while maintaining identical model architectures and training conditions across all comparisons.
We provide the technical experimental details and implementation in Appendix \ref{ap:eval}. 

We report baseline results from various diffusion models; GDSS \citep{gdss} is a continuous diffusion model, DruM \citep{drum} is a diffusion bridge model, and DiGress \citep{digress} is a Discrete Diffusion model.
DiGress uses the marginal distribution as noise. 
It differs from our Marginal DDM only by the network architecture, and hyperparameters. 
Baseline results are reported from \citet{drum}.
Importantly, these baseline results were obtained using 1000 function evaluation steps during generation, whereas we use only 500 steps in our experiments.
We report additional computational cost details for our model and the baselines in Appendix \ref{ap:comp_cost}.

\subsection{Molecule Generation}

We report the Frechet ChemNet Distance (FCD) \citep{fcd}, which measures the similarity between generated molecules and real molecules in chemical space, as well as the Neighborhood Subgraph Pairwise Distance Kernel (NSPDK - NPK) \citep{nspdk}, which evaluates the similarity of their graph structures. Additionally, we report the proportion of chemically valid molecules (validity) without any post-generation correction or resampling. 
\setlength{\tabcolsep}{2pt}
\begin{table}[!ht]
    \centering
    \caption{Molecule generation results on \texttt{QM9} and \texttt{ZINC250k}.} 
    \label{tab:qm9_zinc}
    \begin{center}
    \begin{small}
    \begin{sc}
    \begin{tabular}{lcccccc}
        & \multicolumn{3}{c}{QM9} & \multicolumn{3}{c}{ZINC250k} \\ 
        Model  & Val.$\uparrow$& NPK$\downarrow$& FCD$\downarrow$& Val.$\uparrow$ & NPK$\downarrow$& FCD$\downarrow$ \\ \hline
        GDSS  & $95.72$ & $3.3$ & $2.90$ & $97.01$ & $19.5$ & $14.65$ \\  
        DruM  & $99.69$ & $\mathbf{0.2}$ & $0.11$ & $98.65$ & $\mathbf{1.5}$ & $2.25$ \\  
        DiGress  & $98.19$ & $0.3$ & $\mathbf{0.10}$ & $94.99$ & $2.1$ & $3.48$ \\ 
        Marg. DDM & $95.73$ & $1.92$ & $1.09$ & $80.40$ & $12.96$ & $8.50$ \\ 
        Mask DDM & $48.38$ & $14.75$ & $3.76$ & $8.96$ & $78.63$ & $24.98$ \\  \hline 
        Marg. SID  & $99.67$ & $\mathbf{1.04}$ & $\mathbf{0.50}$ & $99.50$ & $\mathbf{2.06}$ & $\mathbf{2.01}$ \\ 
        Mask SID  & $96.43$ & $1.40$ & $1.80$ & $93.85$ & $11.08$ & $9.05$ \\  
        Mask CID & $\mathbf{99.92}$ & $1.40$ & $1.76$ & $\mathbf{99.97}$ & $2.26$ & $3.46$ \\\hline
    \end{tabular}
    \end{sc}
    \end{small}
    \end{center}
\end{table}
Table~\ref{tab:qm9_zinc} presents the results. The best-performing models among our experiments are bolded. We also bold the best baseline if it outperforms ours.
We generate 10,000 samples, compared against 10,000 test molecules. Results are averaged over five independent sampling runs. Standard deviations and additional metrics, including  uniqueness and novelty, are reported in Appendix \ref{ap:results_comp}.

We observe that our Simple Iterative Denoising (SID) models consistently outperform their Discrete Diffusion Model (DDM) counterparts using identical denoisers on both datasets and this across both noise distributions: marginal and mask.
Furthermore, the results confirm our analysis of compounding denoising error, particularly affecting mask-based discrete diffusion, which performs poorly across both datasets. 
SID and CID address the compounding error successfully. 
Our experiments confirm that the marginal distribution is a more effective noise distribution in graph modeling compared to the mask distribution. 
While compounding denoising errors are less severe with the marginal distribution, SID still consistently outperforms marginal DDM baselines by a wide margin.
Finally, we observe that the Critic further improves the generative performance. 
Notably, on \texttt{ZINC250k}, our Critical Iterative Denoising model reaches a high-level validity rate, producing 50 times less invalid molecules than the best-performing baseline, DruM.

\subsection{Generic Graphs}

For generic graphs, we follow the evaluation protocol introduced by \citet{spectre} and adopted by \citet{gdss} and \citet{drum}, using an 80/20 train–test split with 20\% of the training data reserved for validation.

\begin{table}[!ht]
\caption{Graph generation results on \texttt{Planar} and \texttt{SBM} datasets.}
    \label{tab:planar_sbm}
    \begin{center}
    \begin{small}
    \begin{sc}
    \begin{tabular}{lcccc}
        ~ & \multicolumn{2}{c}{Planar} & \multicolumn{2}{c}
        {SBM} \\ 
        Model & Spect.$\downarrow$ & V.U.N.$\uparrow$ & Spect.$\downarrow$& V.U.N.$\uparrow$ \\ \hline
        GDSS  & $37.0$ & $0$ & $12.8$ & $5$ \\  
        DruM  & $\mathbf{6.2}$ & $90$ & $\mathbf{5.0}$ & $\mathbf{85}$ \\  
        DiGress  & $10.6$ & $75$ & $40.0$ & $74$ \\ 
        Marg. DDM & $83.57$ & $0.0$ & $11.82$ & $0.0$ \\
        Mask DDM & $84.44$ & $0.0$ & $11.38$ & $0.0$ \\  
        \hline
        Marg. SID  & $7.62$ & $\mathbf{91.3}$ & $\mathbf{5.93}$ & $\mathbf{63.5}$ \\
        Mask SID  & $8.72$ & $67.0$ & $15.05$ & $17.5$ \\  
        Mask CID & $\mathbf{6.40}$ & $66.0$ & $11.94$ & $19.0$ \\ \hline
    \end{tabular}
    \end{sc}
    \end{small}
    \end{center}
\end{table}

We evaluate graph similarity using the Spectral Maximum Mean Discrepancy (Spect.), which compares the spectra of the graph Laplacian between generated and real graphs. Unlike other metrics that rely on local structural properties, such as node degrees or clustering coefficients, the Spectral MMD captures graph structures at any level.
Additionally, we report the proportion of Valid, Unique, and Novel graphs (V.U.N.).
For the \texttt{Planar} dataset, a valid graph must be both planar and connected. For the \texttt{Stochastic Block Model} dataset, validity indicates that the generated graph is likely to follow the block model distribution used to generate the training data (i.e., an intra-community edge density of 0.3 and an inter-community edge density of 0.005).
Uniqueness represents the fraction of unique graphs among the generated samples, and novelty the fraction of unique graphs that do not appear in the training set.
In practice, all generated graphs in our experiments were both unique and novel; thus, V.U.N. effectively measures validity.
Table~\ref{tab:planar_sbm} reports the results. We average over five sampling runs, comparing the test set to a generated sample batch of equal size. Standard deviation and  complementary metrics including MMDs based on degree, clustering coefficient, and orbits counts are reported in Appendix \ref{ap:results_comp}.

The results on general graphs align with our findings on molecular graphs, further reinforcing our conclusions.
Our Simple Iterative Denoising models consistently outperform their Discrete Diffusion counterparts, demonstrating their effectiveness across different graph types. Results also confirm that mask-based models are not well suited for graph generation. In particular, on the SBM dataset, their results are barely significant due to their performances.

While the Critic improves spectral MMD on the Planar dataset, its contribution is less pronounced for generic graphs than for molecular graphs. We hypothesize that this is due to deviations from the data distribution being less localized in generic graphs, making it more difficult to discriminate between graph elements.
Using the marginal distribution, our Simple Iterative Denoising model achieves results significantly superior to the Discrete Diffusion baseline.

\subsection{Ablation}

We assess the impact of the Number of Function Evaluations (NFE)—i.e., the number of denoising steps during sampling—on model performance. Specifically, we investigate how the Critic affects the number of steps required to achieve high-quality generation.

We generate graphs with varying NFEs in $\{16,32,64,128,256,512\}$ and evaluate performance on the ZINC250k dataset for Marginal DDM, Marginal SID, and Mask CID. Validity is reported in Figure \ref{fig:nfe_val_init}, while additional plots and full numerical results are provided in Appendix \ref{ap:results_comp}.

Our findings indicate that the Critic substantially reduces the number of denoising steps needed to achieve high validity, Mask CID reaching over 99\% validity in just 32 steps. 
Notably, in the low-NFE regime (16 steps), Mask CID consistently outperforms all other models across all metrics, demonstrating its effectiveness in accelerating inference without sacrificing quality.

\section{Conclusion}

Discrete diffusion models suffer from the Compounding Denoising Error issue, which leads to error accumulation during sampling and negatively affects their performance. We address this issue by introducing an assumption of conditional independence between intermediate noisy states. Thus, the resulting method does not rely on a diffusion process. We further improve upon our method, and introduce a Critic, which steers the distribution of generated elements towards the data distribution, improving sample quality.

Our experimental results demonstrate the effectiveness of our method. Our Simple Iterative Denoising models consistently outperform their corresponding discrete diffusion models. Moreover, our mask model with Critic systematically improves generative performance over the corresponding Simple Iterative Denoising mask model without the Critic.

Additionally, we show that Critical Iterative Denoising significantly reduces the Number of Function Evaluations (NFE) required for sampling, making the generation process more efficient. While this work focuses on graph modeling, future research should explore the applicability of our approach to other structured data domains.

\section*{Acknowledgments}

I am deeply grateful to Prof. Alexandros Kalousis, whose support has been invaluable—far beyond what co-authorship could acknowledge.

I acknowledge the financial support of the Swiss National Science Foundation within the LegoMol project
(grant no. 207428). The computations were performed at the University of Geneva on "Baobab", "Yggdrasil", and "Bamboo" HPC clusters.

\section*{Impact Statement}

This paper presents work whose goal is to advance the field of 
Machine Learning. There are many potential societal consequences 
of our work, none which we feel must be specifically highlighted here.

\bibliography{bib_main}
\bibliographystyle{icml2025}

\newpage
\appendix
\onecolumn
\section{Proofs}\label{ap:proofs}

\subsection{Proposition \ref{prop:forward}: Noising and Diffusion Forward Process}\label{ap:proof_forward}

Given a noise distribution, a data point, and a scheduler, the distribution  $q^{ID}_{t|1}(z|z_1)$  defined by our noising procedure (Equation \ref{eq:noising}) is equal to the distribution $q^{DDM}_{t|1}(z|z_1)$ obtained by the discrete forward process distribution defined in (Equation \ref{eq:stdFwd}). 

\begin{proof}
    We adopt here the notation of \citet{discdiff_austin}, where the probabilities and probability transition are written as matrices, and we start with the discrete time case.

    We have:
    \begin{equation}
        q_{t|t+\Delta_t} = \vx\mQ_t
    \end{equation}

where $\mQ_t$ is a transition probability matrix. In our case, 

    \begin{equation}
        \mQ_t = (1-\beta_t)\mI + \beta\mA,
    \end{equation}

where $\mA$ is an idempotent matrix, representing the noise distribution, the mask distribution being $\mA_{\text{mask}}\vone\ve_m^T$, where $\ve_m$ is the one-hot vector indicating the mask and the marginal distribution being  $\mA_{\text{marg}}\vone\vm^T$, with $\vm$ representing the marginal distribution.

Using the fact that $\mA^2 = \mA$, and denoting $\alpha_t' = 1-\beta$, we observe that:
    \begin{equation}
        \prod_i^\tau \mQ_i = (\prod_i^\tau\alpha_i')\mI + (1 - (\prod_i^\tau\alpha_i'))\mA.
    \end{equation}

With $\alpha_t = \prod_i^\tau\alpha_i'$, we get: $q_t(z|z_1) = \alpha_t \delta_{z_1}(z) + (1-\alpha_t)q_0(z)$.

For the continuous case, we remark that, $\alpha_t$ being smooth, we can turn the product into geometric integral.   
\end{proof}

\subsection{Proposition \ref{prop:corrector}}\label{ap:proofs_corrector}

An \emph{Iterative Denoising} step, described in Equation \ref{eq:denoising}, is equivalent to a corrector sampling discrete diffusion step with a maximal corrector step.

\begin{proof}
    Corrector sampling in discrete diffusion consists of applying $k$ backward (denoising) steps, i.e., predicting $p_{t+k\Delta_t|t}(z|z_t)$ and $k-1$ forward (noising step), $q_{t|t+k\Delta_t}(z|z_{t+(k-1)\Delta_t})$. 
    We call maximal corrector step a corrector sampling step such that $1 - k\Delta_t = t$.
    In this case, a corrector sampling step becomes: 
    
    \begin{align}
         \sum_z p_{1|t}(z|Z_t)q_{t+\Delta_t|1}(z|z_{1}) &= p(z_{t+\Delta_t}|z_{t})\\
         &= \alpha_{t+\Delta_t} p_{1|t}(z | z_t) + (1-\alpha_{t+\Delta_t}) q_0(z),
    \end{align}

    where the second equality follows from Proposition \ref{ap:proofs_denoising}.    
\end{proof}

\subsection{Denoising}\label{ap:proofs_denoising}

Given the random variable $a_t \sim p_{\alpha_t}(a)=\text{Bernoulli}(a; \alpha_t)$ (see Section \ref{sec:cid_prelim}), we have: \(p_{s|1}(z|z_1, a_s) = q_0(z) \: \text{if} \: z=0\), and  \(p_{s|1}(z|z_1, a_s) = \delta_{z_1}(z)\), otherwise. 

\begin{equation}
    \begin{split}
        p_{s|t}(z|z_{t}) &= \sum_{z_1} p_{s|1}(z|z_{t}, z_1) p_{1|t}(z_1|Z_t) \\ 
        &= \sum_{z_1} p_{s|1}(z|z_1) p_{1|t}(z_1|Z_t) \\
        &= \sum_{z_1}\sum_{a} p_{\alpha_s}(a) p_{s|1}(z|z_1, a_s) p_{1|t}(z_1|Z_t) \\
        &= \sum_{z_1} \left(\alpha_s \delta_{z_1}(z) + q_0(z) (1-\alpha_s)\right) p_{1|t}(z_1|Z_t) \\ 
        &= \sum_{z_1} p_{1|t}(z_1|Z_t)  \alpha_s \delta_{z_1}(z) + \sum_{z_1}  p_{1|t}(z_1|Z_t) (1-\alpha_s) q_0(z) \\ 
        &= \alpha_s p_{1|t}(z | Z_t) + (1-\alpha_s) q_0(z)
    \end{split}
\end{equation}

\subsection{Proposition \ref{prop:optimal_critic}: Optimal Critic}

    The optimal Critic is:
    \begin{equation}
         C^*(\hat{z}_{1|t}) = \frac{\alpha_t p_{data}(\hat{z}_{1|t})}{\alpha_tp_{data}(\hat{z}_{1|t}) + (1-\alpha_t)p_{pred}(\hat{z}_{1|t})}
    \end{equation}
    
\begin{proof}
    The proof is inspired by \citet{goodfellow}. 

    For a single element, from Equation \ref{eq:critic_objective}, we have:

    \begin{equation}\label{eq:oc_proof}
    \begin{split}
        \gL_\phi &= - \mathbb{E}_{t, \hat{Z}_{1|t}} \log\left( p_\phi(a|\hat{Z}_{1|t}) \right) \\
        &= - \alpha_t\mathbb{E}_{\hat{Z}_{1|t} \sim p_{data}} \log\left( p_\phi(a|\hat{Z}_{1|t}) \right)
         -(1-\alpha_t ) \mathbb{E}_{\hat{Z}_{1|t} \sim p_{pred}} \log\left( 1-p_\phi(a|\hat{Z}_{1|t}) \right) \\
         &=- \sum_{z_1} \alpha_t p_{data}(\hat{z}_{1|t}) \log\left( p_\phi(a|\hat{Z}_{1|t}) \right) -\sum_{z_1}(1-\alpha_t ) p_{pred}(\hat{z}_{1|t})  \log\left( 1-p_\phi(a|\hat{Z}_{1|t}) \right) \\
            &=- \sum_{z_1} \left( \alpha_t p_{data}(\hat{z}_{1|t}) \log\left( p_\phi(a|\hat{Z}_{1|t}) \right) +(1-\alpha_t ) p_{pred}(\hat{z}_{1|t})  \log\left( 1-p_\phi(a|\hat{Z}_{1|t}) \right) \right)    
    \end{split}
    \end{equation}

    For any $(u, v) \in \mathbb{R}^2\setminus\{(0, 0)\}$ and $y \in [0, 1]$, the function $f(y) = u \text{log}(y) + v\text{log}(1-y)$ reaches its maximum at $\frac{u}{u+v}$. 
    
    Hence, Equation \ref{eq:oc_proof} reaches its minimum at:
        \begin{equation}
         C^*(\hat{z}_{1|t}) = \frac{\alpha_t p_{data}(\hat{z}_{1|t})}{\alpha_tp_{data}(\hat{z}_{1|t}) + (1-\alpha_t)p_{pred}(\hat{z}_{1|t})}
        \end{equation}
\end{proof}

\subsection{Lemma \ref{prop:alphas_compar}: Noising Rate}\label{ap:alpha_compar}
    If $p_{data}(\hat{z}_{1|t}) > p_{pred}(\hat{z}_{1|t})$, the optimal critic's noising rate  
    $\hat{\beta}_t^*$ is smaller that the schedule's noising rate $\beta_t$, that is:
    \begin{equation}
        p_{data}(\hat{z}_{1|t}) > p_{pred}(\hat{z}_{1|t})  \implies \alpha_t < \hat{\alpha}_t, 
    \end{equation}
    and conversely
    \begin{equation}
        p_{data}(\hat{z}_{1|t}) < p_{pred}(\hat{z}_{1|t})  \implies \alpha_t > \hat{\alpha}_t.
    \end{equation}

\begin{proof}
From Theorem \ref{prop:optimal_critic}, we have:
    \begin{align}
        \hat{\alpha}_t &= \frac{\alpha_t p_{data}(\hat{z}_{1|t}) }{\alpha_t p_{data}(\hat{z}_{1|t}) + (1-\alpha_t) p_{pred}(\hat{z}_{1|t})} \\
        \iff \hat{\alpha}_t^{-1} \alpha_t p_{data}(\hat{z}_{1|t}) &= \alpha_t p_{data}(\hat{z}_{1|t}) + (1-\alpha_t) p_{pred}(\hat{z}_{1|t}) \\
        \iff \hat{\alpha}_t^{-1} p_{data}(\hat{z}_{1|t}) &= p_{data}(\hat{z}_{1|t}) + (\alpha_t^{-1}-1) p_{pred}(\hat{z}_{1|t}) \\
        \iff \frac{p_{data}(\hat{z}_{1|t})}{p_{pred}(\hat{z}_{1|t})} &= \frac{\alpha_t^{-1}-1}{\hat{\alpha}_t^{-1}-1}
    \end{align}
    
    Hence:
    \begin{equation}
        p_{data}(\hat{z}_{1|t}) > p_{pred}(\hat{z}_{1|t}) \implies \frac{\alpha_t^{-1}-1}{\hat{\alpha}_t^{-1}-1} > 1 \implies \alpha_t < \hat{\alpha}_t
    \end{equation}
    
    \begin{equation}
        p_{data}(\hat{z}_{1|t}) < p_{pred}(\hat{z}_{1|t})  \implies \alpha_t > \hat{\alpha}_t
    \end{equation}
\end{proof}

\section{Comparing Simple Iterative Denoising with Discrete Diffusion and Discrete Flow Matching}

In Table \ref{tab:compar_summary}, we compare Simple Iterative Denoising with Discrete Diffusion and Discrete Flow Matching.
For Discrete Diffusion, we follow the framework of \citet{discdiff_austin}, which uses uniform and mask (absorbing state) noise , and its extension to the marginal setting by \citet{digress}.
For Discrete Flow Matching, we adopt the standard formulation from \citet{discrete_flow_matching_gat}, which uses independent coupling and convex interpolants.
We use the following shorthand notations: $s = t+\Delta_t$ and $D_t = \Delta_t\frac{\dot{\alpha_t}}{1 -\alpha_t}$. 

\setlength{\tabcolsep}{3pt}
\renewcommand{\arraystretch}{1.8}
\begin{table}[!ht]
    \caption{Comparing Simple Iterative Denoising, Discrete Diffusion, and Discrete Flow Matching.}
    \label{tab:compar_summary}
    \begin{center}
    \begin{sc}
    \begin{tabular}{|l|c|c|c|}
        \hline
        & \multicolumn{1}{c|}{Simple Iterative Denoising} 
        & \multicolumn{1}{c|}{Discrete Diffusion} 
        & \multicolumn{1}{c|}{Discrete Flow Matching} \\ 
        \hline
        Noising
        & Conditional indep. given $z_1$ 
        & Markovian 
        & Markovian \\ 
        \hline  
        $p_{t|s, 1}(z|z_s, z_1)$  
        & $p_{t|1}(z|z_1)$ 
        & $p_{t|s}(z|z_s)$ 
        & $p_{t|s}(z|z_s)$
        \\ \hline
        $q_{t|1}(z|z_1)$ 
        & $\alpha_t \delta_{z_1}(z) + (1-\alpha_t)q_0(z)$ 
        & $\alpha_t \delta_{z_1}(z) + (1-\alpha_t)q_0(z)$ 
        & $\alpha_t \delta_{z_1}(z) + (1-\alpha_t)q_0(z)$ 
        \\ \hline
        $q_{s|t}(z |Z_t, z_1)$ 
        & $q_{s|1}(z |z_1)$ 
        & $\frac{q(z_t \mid z_s) \, q(z_s \mid z_1)}{q(z_t \mid z_1)}$ 
        & $(1 - D_t)\delta_{z_t}(z) + D_t \delta_{z_1}(z)$ 
        \\ \hline
        $p_{s|t}(z|z_t)$  
        & $\alpha_s p_{1|t}(z \mid Z_t) + (1-\alpha_s) q_0(z)$ 
        & $\sum_{z_1} q_{s|t}(z |Z_t, z_1)p_{1|t}(z \mid Z_t)$ 
        & $\delta_{z_t}(z) + D_t \left( p_{1|t}(z \mid Z_t) - \delta_{z_t}(z) \right)$
        \\ \hline      
    \end{tabular}
    \end{sc}
    \end{center}
\end{table}
\setlength{\tabcolsep}{3pt}
\renewcommand{\arraystretch}{1}

\begin{figure}[H]
    \centering
    \caption{Noising and Denoising Processes in Discrete Diffusion and Simple Iterative Denoising}
    \subfigure[Mask Discrete Diffusion]{
        \includegraphics[width=0.15\textwidth]{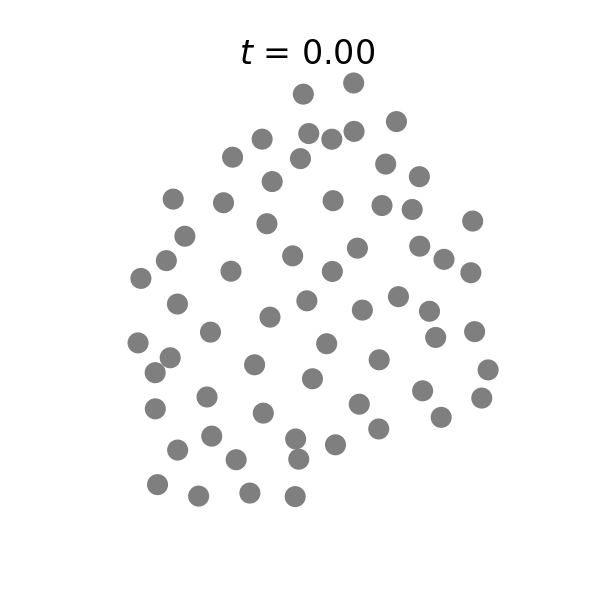}
        \includegraphics[width=0.15\textwidth]{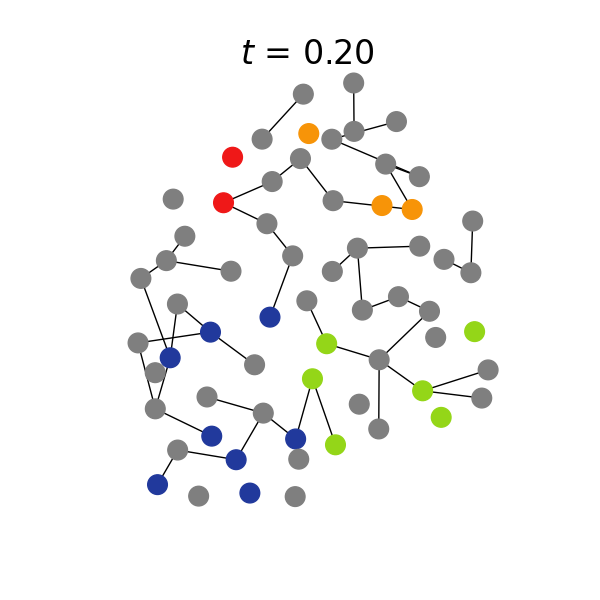}
        \includegraphics[width=0.15\textwidth]{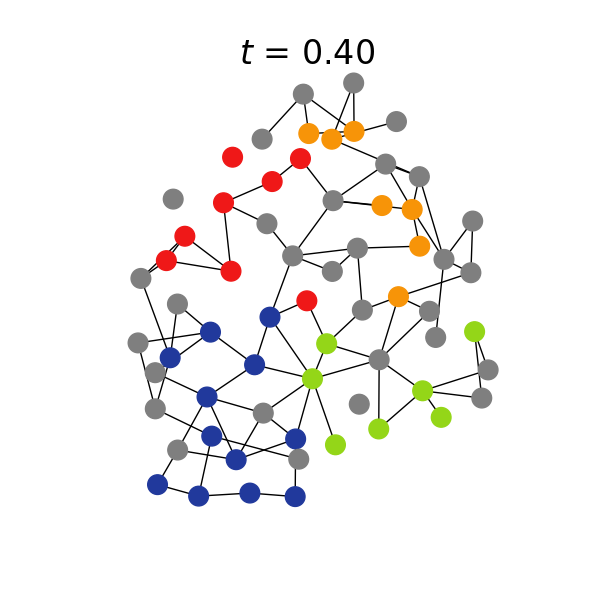} 
        \includegraphics[width=0.15\textwidth]{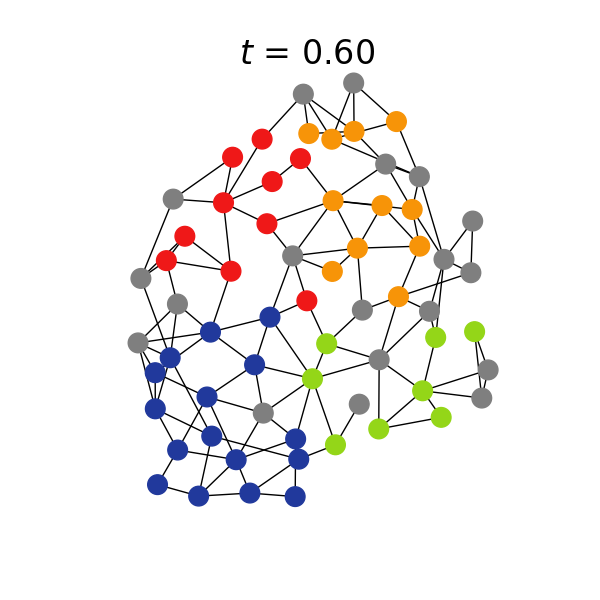} 
        \includegraphics[width=0.15\textwidth]{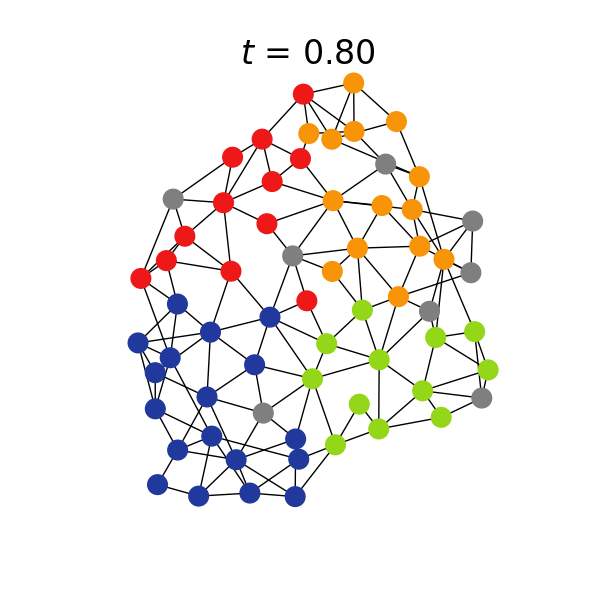} 
        \includegraphics[width=0.15\textwidth]{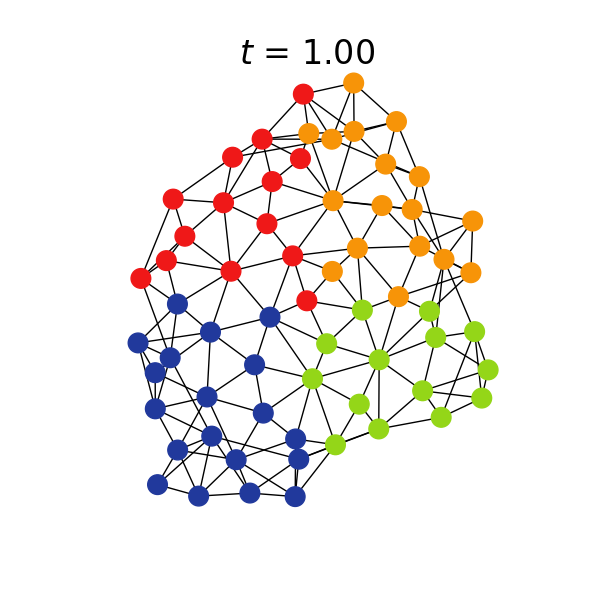}
    }
        \subfigure[Mask Simple Iterative Denoising]{
        \includegraphics[width=0.15\textwidth]{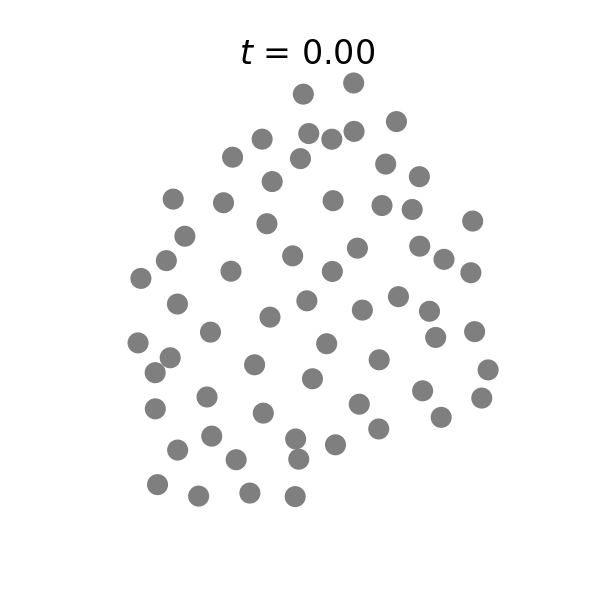}
        \includegraphics[width=0.15\textwidth]{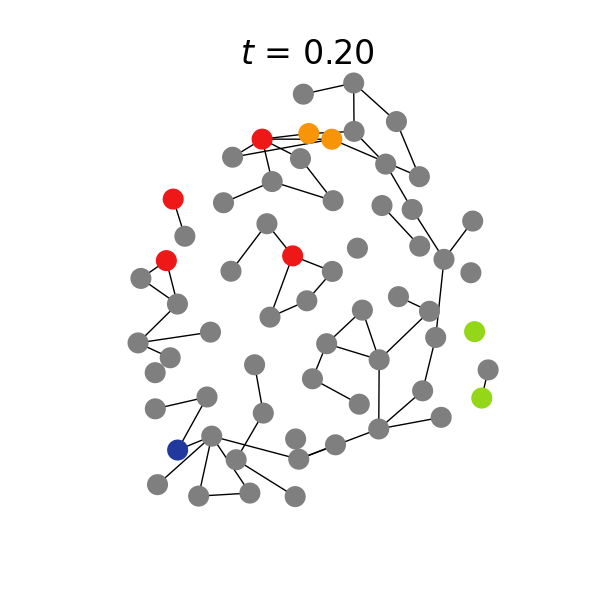}
        \includegraphics[width=0.15\textwidth]{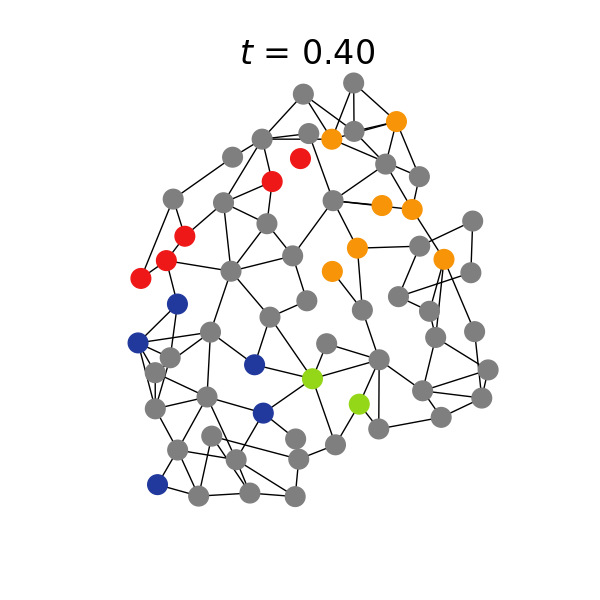} 
        \includegraphics[width=0.15\textwidth]{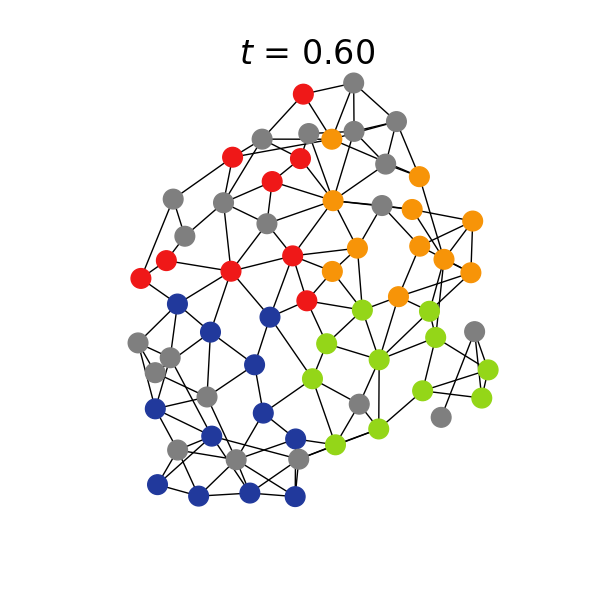} 
        \includegraphics[width=0.15\textwidth]{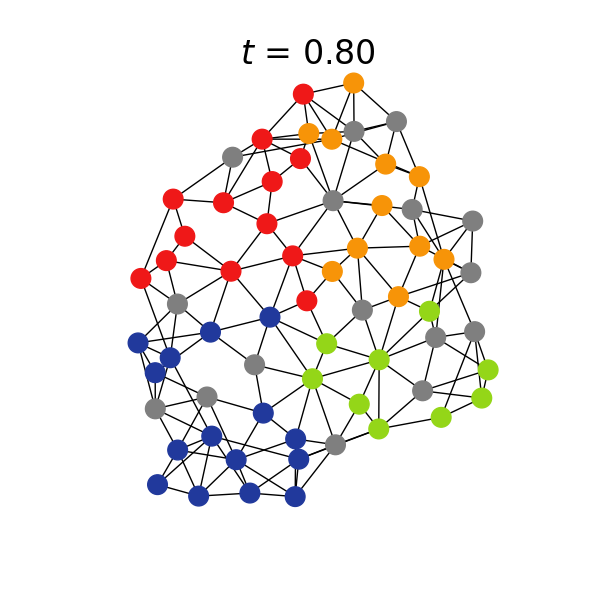} 
        \includegraphics[width=0.15\textwidth]{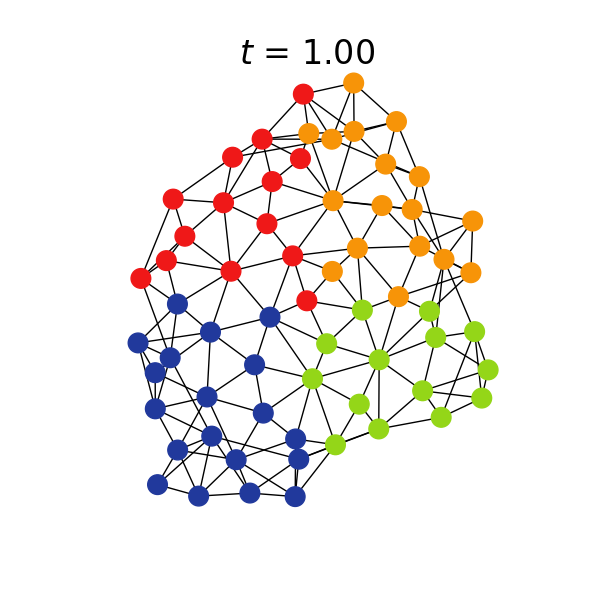}
    }
\end{figure}

Unlike standard Discrete Diffusion, in Simple Iterative Denoising, an element that is masked (in forward) or unmasked (in backward) can be unmasked (in forward) or remasked (in backward). Note that for the sake of illustration, we use the absence of edge as an absorbing state. In practice, we use three different states for mask SID and mask DDM (mask, edge, non-edge).

\section{Generative Graph Modeling: Related Works}\label{ap:related_works}

A main challenge in generative graph modeling follows from the $n!$ different ways to represent graphs due to node permutation. This has motivated two dominant families of approaches: \textit{sequential models}, which generate graphs step-by-step, and \textit{equivariant models}, which preserve equivariance to node permutation by design.

\paragraph{Sequential models} construct graphs auto-regressively, adding nodes, edges, or substructures in a predefined order \citep{graphrnn, graphaf, graphdf, gran, grapharm, pard}. To limit the number of possible sequences representing a single graph, many methods adopt a Breadth-First Search (BFS) traversal strategy. In specific domains such as molecular graph generation, canonical formats like SMILES have been used to mitigate permutation issues \citep{gomez-bombarelli_automatic_2018, kusner_grammar_2017}. However, general-purpose canonization strategies \citep{graphgen} fail for large graphs (see experiments in \citet{graphle}). Hierarchical models that aggregate subgraphs \citep{jtvae, hiervae} also fall into this category, though they rely on enumerating a predefined vocabulary of substructures and are thus limited to constrained domains.

\paragraph{Equivariant models} aim to preserve permutation symmetry by producing the same output regardless of node ordering. These models have been instantiated across several generative paradigms, including GANs \citep{gggan, spectre}, normalizing flows \citep{graphnvp, moflow, gnf}, and vector-quantized auto-encoders \citep{dgae, glad}. More recently, equivariant denoising models in continuous spaces as score-based diffusion \citep{edp-gnn, gdss}, diffusion bridges \citep{drum}, and Flow Matching \citep{VarFlowMatching} have significantly improved graph generation for small graphs. Despite their success, these continuous methods introduce a mismatch between the continuous noise space and the inherently discrete structure of graphs.

To address this, several works have explored discrete denoising processes. Discrete diffusion models in both discrete time \citep{discdiff_haefeli, digress} and continuous time \citep{continuous_time_graph_diffusion} have demonstrated strong performance in generating small graphs. Discrete flow matching has also been adapted for graphs, with \citet{DeFog} showing promising results. However, these discrete models suffer from compounding denoising errors. Our proposed method directly tackles this issue, leading to improved generative performance.

\paragraph{Scalability.} A key limitation of current equivariant models is their reliance on dense representations and pairwise computations, which hinder scalability to large graphs. To address this, recent works have proposed scalable denoising architectures and sampling strategies \citep{sparsediff, EDGE, higen, graphle, head}. 
Notably, \citet{sparsediff} and \citet{head} propose methods that enable any equivariant model to scale to large graphs. While these methods could extend our iterative denoising framework to larger graphs, we leave this exploration for future research.

\section{Models}

\subsection{GNNs architecture}

Our denoisers are Graph Neural Network, inspired by the general, powerful, scalable (GPS) graph Transformer.

We implement a single layer as: 

\begin{align}
    \tilde{\mX}^{(l)}, \tilde{\mE}^{(l)} &= \text{MPNN}(\mX^{(l)}, \mE^{(l)}), \\
    \mX^{(l+1)} &= \text{MultiheadAttention}({\tilde{\mX}^{(l)}} + \mX^{(l)}) + {\tilde{\mX}^{(l)}} \\
    \mE^{(l+1)} &= \tilde{\mE}^{(l)} + \mE^{(l)}
\end{align}

where, $\mX^{(l)}$ and $\mE^{(l)}$ are the node and edge hidden representations after the $l$\textsuperscript{th} layer. The \emph{Multihead Attention }layer is the classical multi-head attention layer from \citet{attention_is_all}, and MPNN is a Message-Passing Neural Network layer described hereafter. 

The MPNN operates on each node and edge representations as follow:

\begin{align}
    \vh^{l}_{i, j} &= \text{ReLU}(\mW^{l}_{src}\vx^{l}_i + \mW^{l}_{trg}\vx^{l}_j + \mW^{l}_{edge}\ve^{l}_{i, j}) \\
    \ve^{l+1}_{i, j} &= \text{LayerNorm}(f_{\text{edge}}(\vh^{l}_{i, j}) \\
 	\vx^{l+1}_{i} &= \text{LayerNorm}\left(\vx^{l}_{i} + \sum_{j \in \mathcal{N}(i)}f_{\text{node}}(\vh^{l}_{i, j})\right),  
\end{align}

where $\mW^{l}_{src}$, $\mW^{l}_{trg}$, and $\mW^{l}_{edge}$ are matrices of parameters and $f_{\text{node}}$, and $f_{\text{edge}}$ are small neural networks.

The node hidden representation, i.e., the $\vx_i$'s and the hidden representation of $f_{\text{node}}$ are of dimensions $d_h$, an hyperparameter (see Table \ref{tab:hyperparam_denoising_fix}). 
The edge hidden representation, i.e., the $\ve_{i, j}$'s, $\vh^{l}_{i, j}$'s, and the hidden representation of $f_{\text{edge}}$ are of dimensions $d_h/4$. 

\paragraph{Inputs and Outputs}

In the input, we concatenate the node attributes, extra features, and time step as node features, copying graph-level information (e.g., time step or graph size) to each node. The node and edge input vectors are then projected to their respective hidden dimensions, \(d_h\) for nodes and \(d_h/4\) for edges.

Similarly, the outputs of the final layer are projected to their respective dimensions, \(d_x\) for nodes and \(d_e\) for edges (or to a scalar in the case of the \emph{Critic}). To enforce edge symmetry, we compute 
$\ve_{i, j} = \frac{\ve_{i, j} + \ve_{j, i}}{2}.$
Finally, we ensure the outputs can be interpreted as probabilities by applying either a softmax or sigmoid function, as appropriate.

\subsection{hyperparameters}

As explained here above, the node hidden representation has size $d_h$ and the edge representation $d_h/4$. 
We use $d_h=64$ with the \texttt{QM9} dataset and $d_h=256$ with all the other datasets. 

\begin{table}[h]
    \centering
    \caption{Hyperparameters}
    \begin{tabular}{ll}
        MPNN layers   & 4 \\
        Layers in MLPs   & 3 \\
        Diffusion steps & 500 \\
        Learning rate &  0.0002 \\
        Optimizer & Adam \\ 
        Betas parameters for Adam & (0.9, 0.999) \\ 
        Scheduler & cosine \citep{cosine_sched}
    \end{tabular}
    \label{tab:hyperparam_denoising_fix}
\end{table}

\subsection{Extra Features}

Following a common practice \citep{digress, DeFog, head}, we enhance the graph representation with synthetic extra node features. We use the following extra features: eigen features, graph size, molecular features (for molecular datasets such as \texttt{QM9} and \texttt{ZINC250k}), and cycle information (for the \texttt{Planar} datasets). All these features are concatenated to the input node attributes.

\paragraph{Spectral features}
We use the eigenvectors associated with the \(k\) lowest eigenvalues of the graph Laplacian. Additionally, we concatenate the corresponding \(k\) lowest eigenvalues to each node.

\paragraph{Graph size encoding}
The graph size is encoded as the ratio between the size of the current graph and the largest graph in the dataset, \(n/n_{\text{max}}\). This value is concatenated to all nodes in the graph.

\paragraph{Molecular features}
For molecular datasets, we use the charge and valency of each atom as additional features.

\paragraph{Cycles}
Following \citet{digress}, we count the number of cycles of size 3, 4, and 5 that each node is part of, and use these counts as features.

\section{Evaluation}\label{ap:eval}
\subsection{Molecular Benchmark}

For molecular graphs, we adopt the evaluation procedure followed by \citet{drum}, from which we took the baseline model results, and which was originally established in \citet{gdss}.

\paragraph{Datasets}

The QM9 dataset \citep{qm9} consists of 133,885 organic molecules with up to 9 heavy atoms, including carbon (C), oxygen (O), nitrogen (N), and fluorine (F). In contrast, the ZINC250k dataset \citep{zinc} contains 250,000 molecules with up to 38 atoms spanning 9 element types: C, O, N, F, phosphorus (P), sulfur (S), chlorine (Cl), bromine (Br), and iodine (I). Both datasets are divided into a test set (25,000 molecules), a validation set (25,000 molecules), and a training set (the remaining molecules).

For our experiments, we preprocess the datasets following standard procedures \citep{gdss, drum}. Molecules are kekulized using RDKit, and explicit hydrogen atoms are removed from the QM9 and ZINC250k datasets.

We evaluate the models using three metrics:
\begin{enumerate}
    \item Validity: The percentage of chemically valid molecules among the generated samples, determined using RDKit's \texttt{Sanitize} function without applying post-hoc corrections, such as valency adjustments or edge resampling.
    \item Fréchet ChemNet Distance (FCD) \citep{fcd}: Measures the distance between feature distributions of generated molecules and test set molecules, using ChemNet to capture their chemical properties.
    \item Neighborhood Subgraph Pairwise Distance Kernel (NSPDK) MMD \citep{nspdk}: Assesses the quality of graph structures by computing the maximum mean discrepancy (MMD) between the generated molecular graphs and those from the test set.
\end{enumerate}

\subsection{Generic Graphs}

We assess the quality of the generated graphs using three benchmark datasets from \citet{spectre}.

Planar Graph Dataset: This dataset consists of 200 synthetic planar graphs, each containing 64 nodes. A generated graph is considered valid if it is connected and planar.

Stochastic Block Model (SBM) Dataset: This dataset consists of 200 synthetic graphs generated using the Stochastic Block Model (SBM). The number of communities is randomly sampled between 2 and 5, and the number of nodes per community is sampled between 20 and 40. The edge connection probabilities are set as follows:
\begin{itemize}
    \item Intra-community edges: 0.3 (probability of an edge existing within the same community).
    \item Inter-community edges: 0.005 (probability of an edge existing between different communities).
\end{itemize}        
A generated graph is considered valid if it satisfies the statistical test introduced in \cite{spectre}, which assess it corresponds to to the SBM structure.

We adopt the evaluation framework proposed by \cite{gran}, utilizing total variation (TV) distance to measure the Maximum Mean Discrepancy (MMD). This approach is significantly more computationally efficient than using the Earth Mover’s Distance (EMD) kernel, particularly for large graphs.

Additionally, we employ the V.U.N. metric as introduced by Martinkus et al. (2022), which evaluates the proportion of valid, unique, and novel graphs among the generated samples. A graph is considered valid if it satisfies the dataset-specific structural properties described earlier. 

\subsection{Computational Cost and Baselines}\label{ap:comp_cost}

\subsubsection{Additional Overhead from CID}

CID introduces a computational overhead, as it requires training a Critic and evaluating it at each denoising step. As a first-order approximation, this doubles both training and inference time. However, the performance gain over Mask ID and Mask DDM is important, and Mask denoising models are suitable for some specific downstream applications such as molecular scaffold extension.

\subsubsection{Comparison with Baselines}

For both our model and baseline methods, the dominant computational cost arises from the architecture—especially from the size of edge feature vectors, on which MLPs are applied, as their number scales quadratically with the number of nodes in the graph, and linearly with the number of layers.

Below is a comparison based on DruM’s configuration files, summarizing the number of layers ($L$) and edge feature dimensions ($d_E$) per dataset:

\begin{table}[h]
\centering
\caption{Number of layers ($L$) and hidden dimension ($d_E$) used for our models and DruM across datasets.}
\begin{sc}
\begin{tabular}{lcc}
\textbf{Dataset} & \textbf{Ours ($L \times d_E$)} & \textbf{DruM ($L \times d_E$)} \\
\midrule
Qm9     & 4$\times$16   & 8$\times$256 \\
Zinc    & 4$\times$64   & 9$\times$128 \\
Planar  & 4$\times$64   & 8$\times$64  \\
SBM     & 4$\times$64   & 8$\times$64  \\
\bottomrule
\end{tabular}
\end{sc}
\end{table}

We highlight that we evaluate our model using 500 denoising steps, whereas the baselines use 1000 steps.

In summary, our model achieves superior performance in key practical tasks, such as large molecule generation (as measured by validity and Fréchet ChemNet Distance), using fewer denoising steps, fewer layers, and smaller per-layer costs. We appreciate the reviewer’s remark to make and will update the manuscript in order to explicit these advantages.
Evidence Supporting the Effectiveness of Our Method

Our work addresses a core limitation of discrete denoising models. We show that our approach consistently outperforms existing discrete diffusion methods across all evaluation metrics, including those ablating NFE's. On this regard, we believe the paper provides sufficient evidence of our method’s effectiveness.

\section{Complementary Experimental Results}\label{ap:results_comp}    


\paragraph{Results on \texttt{QM9}}

As additional metrics, we report uniqueness, defined as the fraction of unique molecules among the generated samples, and novelty, the fraction of unique molecules that are not present in the training dataset.
Furthermore, all models achieve 100\% validity with valency correction.

\begin{table}[H]
    \centering
    \caption{Results on \texttt{QM9}}
    \begin{sc}
    \begin{tabular}{lccccc}
        Average & Valid & Unique & Novel & NSPDK & FCD \\ \hline
        Marginal DDM & $95,73 \pm 0,24$ & $97,52 \pm 0,16$ & $79,55 \pm 0,28$ & $1,922 \pm 0,054$ & $1,090 \pm 0,035$ \\
        Marginal ID & $99,67 \pm 0,06$ & $95,66 \pm 0,28$ & $72,57 \pm 0,52$ & $1,041 \pm 0,049$ & $0,504 \pm 0,010$ \\ 
        Mask DDM & $48,38 \pm 0,47$ & $78,47 \pm 0,65$ & $75,21 \pm 0,88$ & $14,750 \pm 0,509$ & $3,760 \pm 0,035$ \\
        Mask ID & $96,43 \pm 0,16$ & $98,02 \pm 0,14$ & $86,79 \pm 0,31$ & $1,395 \pm 0,015$ & $1,797 \pm 0,027$ \\
        Mask CID & $99,92 \pm 0,02$ & $96,93 \pm 0,13$ & $80,74 \pm 0,26$ & $1,402 \pm 0,034$ & $1,757 \pm 0,035$ \\ 
    \end{tabular}
    \end{sc}
\end{table}

\paragraph{Results on \texttt{ZINC250k}}

As additional metrics, we report uniqueness, defined as the fraction of unique molecules among the generated samples, and novelty, the fraction of unique molecules that are not present in the training dataset.
Furthermore, all models achieve 100\% validity with valency correction.

\begin{table}[H]
    \centering
    \caption{Results on \texttt{ZINC250k}}
        \begin{sc}
    \begin{tabular}{lccccc}
        ~ & Valid & Unique & Novel & NSPDK & FCD \\ \hline
        Marginal DDM & $80,40 \pm 0,31$ & $95,93 \pm 0,02$ & $99,98 \pm 0,03$ & $12,96 \pm 0,25$ & $8,50 \pm 0,09$ \\ 
        Marginal ID & $99,50 \pm 0,06$ & $99,84 \pm 0,02$ & $99,97 \pm 0,00$ & $2,06 \pm 0,05$ & $2,01 \pm 0,01$ \\ 
        Mask DDM & $8,96 \pm 0,36$ & $99,37 \pm 0,00$ & $100,00 \pm 0,24$ & $78,63 \pm 2,41$ & $24,98 \pm 0,24$ \\ 
        Mask ID & $93,85 \pm 0,25$ & $100,00 \pm 0,01$ & $100,00 \pm 0,02$ & $11,08 \pm 0,17$ & $9,05 \pm 0,16$ \\ 
        Mask CID & $99,97 \pm 0,01$ & $99,98 \pm 0,01$ & $99,98 \pm 0,01$ & $2,26 \pm 0,09$ & $3,46 \pm 0,01$ \\ 
    \end{tabular}
    \end{sc}
\end{table}

\paragraph{Results on \texttt{Planar}}

We provide additionally the results using the degree, clustering and orbit MMD. 

\begin{table}[H]
    \centering
    \caption{Results on \texttt{Planar}}
        \begin{sc}
    \begin{tabular}{lccccc}
        ~ & Spectral & V.U.N. & Degree & Clustering & Orbit \\ \hline
        Marginal DDM & $83,57 \pm 2,56$ & $0,0 \pm 0,0$ & $53,83 \pm 0,80$ & $300,32 \pm 3,15$ & $1441,45 \pm 83,57$ \\ 
        Marginal ID & $7,62 \pm 1,34$ & $91,3 \pm 4,1$ & $5,93 \pm 1,26$ & $163,40 \pm 31,86$ & $19,08 \pm 4,14$ \\ 
        
         Mask DDM & $84,44 \pm 2,72$ & $0,0 \pm 0,0$ & $57,07 \pm 2,15$ & $297,75 \pm 2,50$ & $1397,94 \pm 34,17$ \\ 
        Mask ID & $8,72 \pm 1,37$ & $67,0 \pm 6,5$ & $2,30 \pm 0,67$ & $78,48 \pm 13,12$ & $11,68 \pm 4,78$ \\ 
        Mask CID & $6,40 \pm 1,20$ & $66,0 \pm 5,5$ & $2,11 \pm 0,62$ & $85,86 \pm 10,74$ & $14,18 \pm 5,54$ \\ 
        
    \end{tabular}
    \end{sc}
\end{table}

\paragraph{Results on \texttt{SBM}}

We provide additionally the results using the degree, clustering and orbit MMD. 

\begin{table}[H]
    \centering
    \caption{Results on \texttt{SBM}}
        \begin{sc}
    \begin{tabular}{lccccc}
        ~ & Spectral & V.U.N. & Degree & Clustering & Orbit \\ \hline
        Marginal DDM & $11,82 \pm 0,85$ & $0,0 \pm 0,0$ & $0,96 \pm 0,73$ & $85,66 \pm 6,26$ & $72,37 \pm 5,51$ \\ 
        Marginal ID & $5,93 \pm 1,18$ & $63,5 \pm 3,7$ & $11,54 \pm 2,72$ & $51,41 \pm 1,49$ & $123,14 \pm 5,35$ \\ 
        Mask DDM & $11,38 \pm 1,04$ & $0,0 \pm 0,0$ & $3,81 \pm 1,79$ & $83,13 \pm 2,27$ & $123,00 \pm 3,73$ \\ 
        Mask ID & $15,05 \pm 4,29$ & $17,5 \pm 5,7$ & $70,12 \pm 21,99$ & $55,63 \pm 0,66$ & $127,05 \pm 17,95$ \\
        Mask CID & $11,94 \pm 2,71$ & $19,0 \pm 4,9$ & $43,13 \pm 12,31$ & $55,06 \pm 2,29$ & $92,80 \pm 16,96$ \\ 
    \end{tabular}
    \end{sc}
\end{table}

\paragraph{NFE Ablation on \texttt{ZINC250}}

We present here addition figures and tables ablating the effect of the NFE.

\begin{figure}[H]
    \label{fig:nfe}
    \begin{center}
     \includegraphics[scale=0.25]{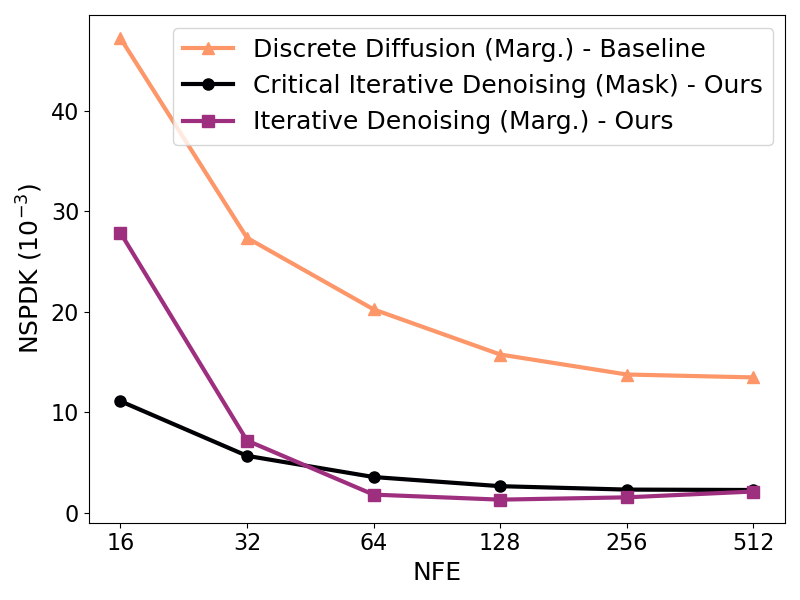}
     \includegraphics[scale=0.25]{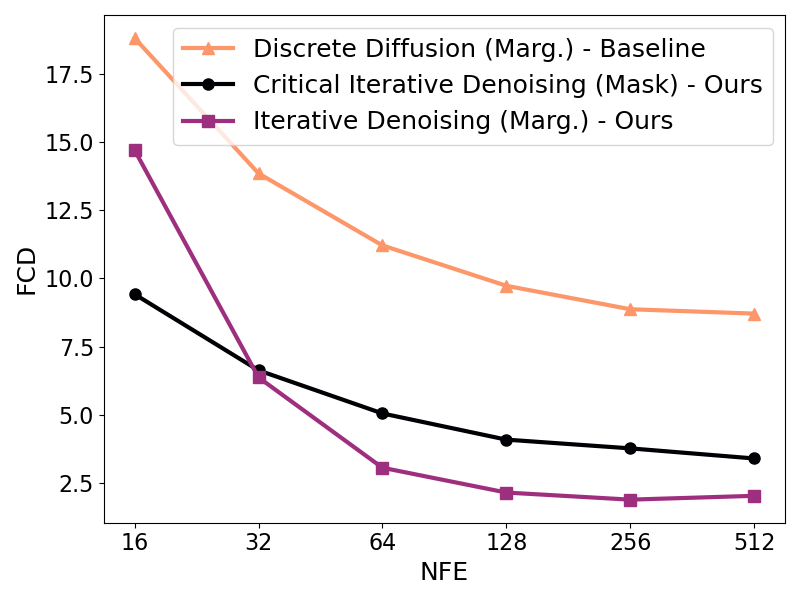}
    \end{center}
    \caption{NSPDK and FCD as a function of the Number of Function Evaluations (NFE) for three models: Discrete Diffusion (baseline), Iterative Denoising (ours), and Critical Iterative Denoising (ours).}
\end{figure}

\begin{table}[!ht]
    \centering
    \caption{Validity vs NFE}
    \begin{sc}
    \begin{tabular}{lcccccc}
        NFE & 16 & 32 & 64 & 128 & 256 & 512 \\ \hline
        Mask ID & 36,34 & 56,60 & 70,94 & 81,78 & 89,37 & 93,57 \\ 
        Mask DDM & 9,35 & 8,84 & 9,05 & 9,14 & 9,06 & 8,49 \\ 
        Mask CID & 92,51 & 99,17 & 99,78 & 99,91 & 99,98 & 99,96 \\ 
        Marg. DDM & 57,31 & 69,33 & 76,80 & 78,09 & 80,15 & 80,37 \\ 
        Marg. ID & 38,25 & 73,66 & 92,71 & 97,69 & 99,06 & 99,58 \\ 
    \end{tabular}
    \end{sc}
\end{table}

\begin{table}[H]
    \centering
    \caption{NSPDK vs NFE}
        \begin{sc}
    \begin{tabular}{lcccccc}
        NFE & 16 & 32 & 64 & 128 & 256 & 512 \\ \hline
        Mask ID & 30,05 & 24,90 & 20,03 & 15,68 & 13,10 & 10,49 \\ 
        Mask DDM & 78,27 & 79,66 & 77,94 & 79,59 & 79,11 & 77,97 \\ 
        Mask CID & 11,10 & 5,66 & 3,55 & 2,64 & 2,30 & 2,26 \\ 
        Marg. DDM & 47,22 & 27,35 & 20,23 & 15,74 & 13,75 & 13,46 \\ 
        Marg. ID & 27,86 & 7,18 & 1,80 & 1,30 & 1,53 & 2,11 \\ 
    \end{tabular}
    \end{sc}
\end{table}

\begin{table}[H]
    \centering
    \caption{FCD vs NFE}
        \begin{sc}
    \begin{tabular}{lcccccc}
        NFE & 16 & 32 & 64 & 128 & 256 & 512 \\ \hline   
        Marg, DDM & 18,81 & 13,85 & 11,21 & 9,73 & 8,87 & 8,71 \\ 
        Marg ID & 14,70 & 6,37 & 3,06 & 2,15 & 1,89 & 2,03 \\ 
        Mask DDM & 25,23 & 25,35 & 24,73 & 25,09 & 24,96 & 24,69 \\ 
        Mask ID & 15,80 & 14,16 & 12,43 & 10,80 & 9,80 & 9,01 \\ 
        Mask CID & 9,42 & 6,63 & 5,05 & 4,09 & 3,77 & 3,40 \\ 
    \end{tabular}
    \end{sc}
\end{table}

\newpage

\section{Visualizations}\label{ap:visual}
\subsection{Molecular graphs}
\textcolor{white}{-}

\begin{figure}[H]
    \centering
    \caption{QM9}
    \begin{tabular}{|ccc|ccc|}
    \hline
      \multicolumn{3}{|c|}{Generated molecules}   &
      \multicolumn{3}{|c|}{Real molecules}
     \\
    \hline
           
        \includegraphics[width=0.12\textwidth]{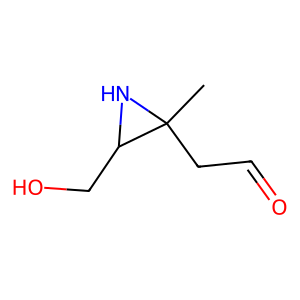} &
        \includegraphics[width=0.12\textwidth]{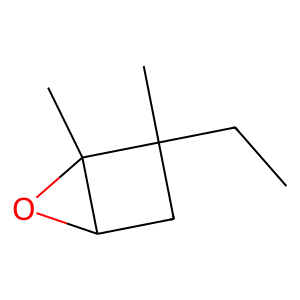} &
        \includegraphics[width=0.12\textwidth]{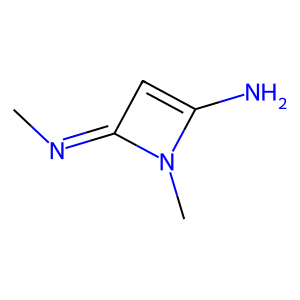} &
        \includegraphics[width=0.12\textwidth]{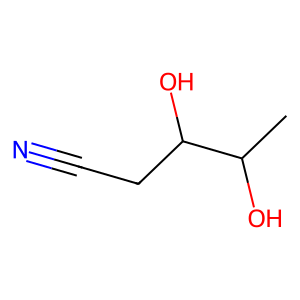} &
        \includegraphics[width=0.12\textwidth]{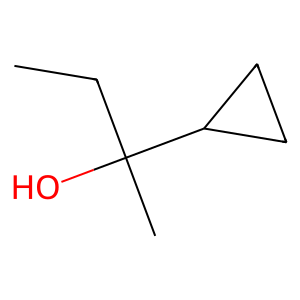}&
        \includegraphics[width=0.12\textwidth]{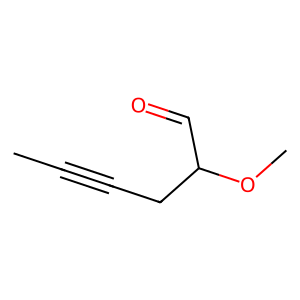} 
        \\
        \includegraphics[width=0.12\textwidth]{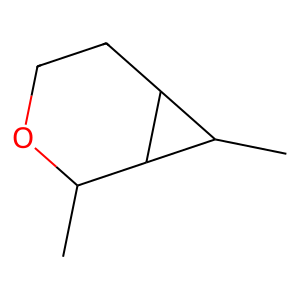} &
        \includegraphics[width=0.12\textwidth]{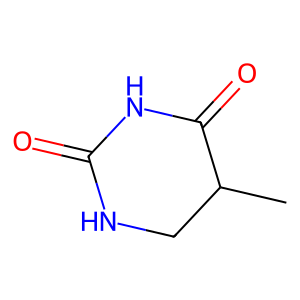} &
        \includegraphics[width=0.12\textwidth]{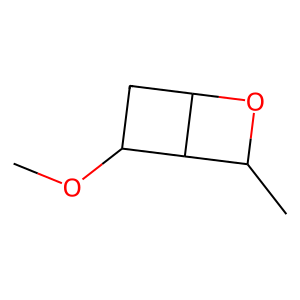} &
        \includegraphics[width=0.12\textwidth]{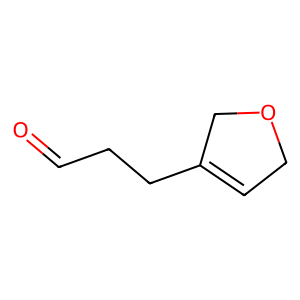} &
        \includegraphics[width=0.12\textwidth]{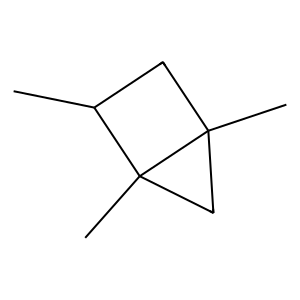}&
        \includegraphics[width=0.12\textwidth]{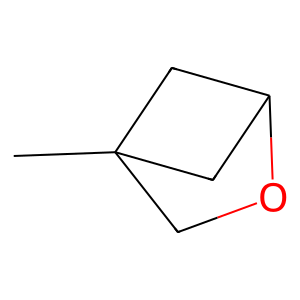} 
        \\
        \includegraphics[width=0.12\textwidth]{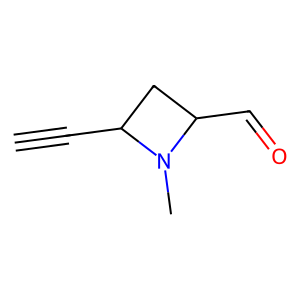} &
        \includegraphics[width=0.12\textwidth]{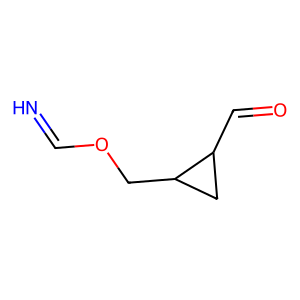} &
        \includegraphics[width=0.12\textwidth]{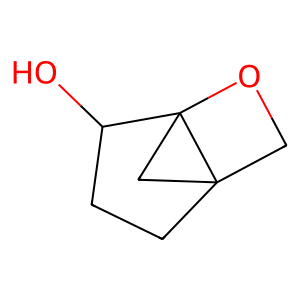} &
        \includegraphics[width=0.12\textwidth]{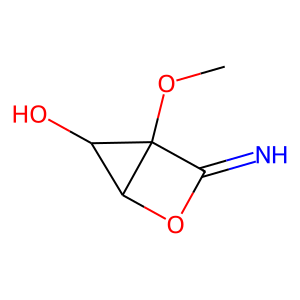} &
        \includegraphics[width=0.12\textwidth]{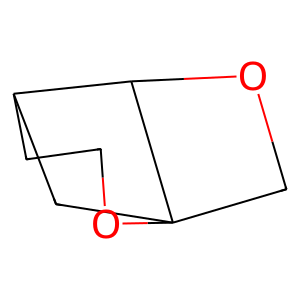}&
        \includegraphics[width=0.12\textwidth]{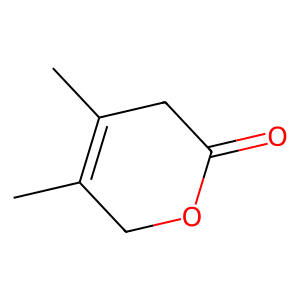} 
        \\
        \includegraphics[width=0.12\textwidth]{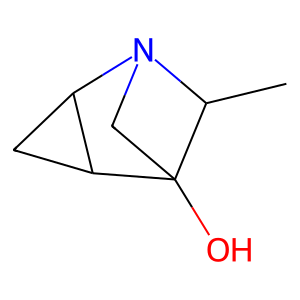} &
        \includegraphics[width=0.12\textwidth]{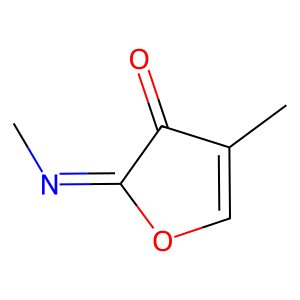} &
        \includegraphics[width=0.12\textwidth]{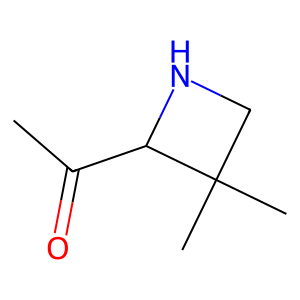} &
        \includegraphics[width=0.12\textwidth]{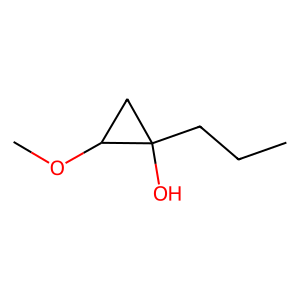} &
        \includegraphics[width=0.12\textwidth]{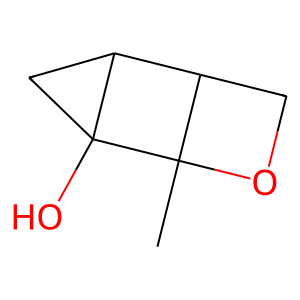}&
        \includegraphics[width=0.12\textwidth]{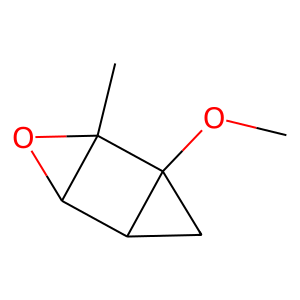} 
        \\
        \hline
        \end{tabular}
\end{figure}

\begin{figure}[H]
    \centering
    \caption{ZINC250K}
    \begin{tabular}{|ccc|ccc|}
    \hline
      \multicolumn{3}{|c|}{Generated molecules}   &
      \multicolumn{3}{|c|}{Real molecules}
     \\
    \hline
           
        \includegraphics[width=0.12\textwidth]{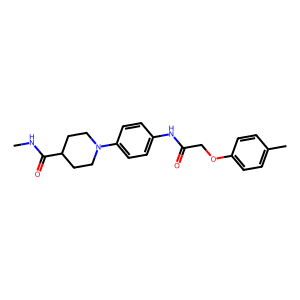} &
        \includegraphics[width=0.12\textwidth]{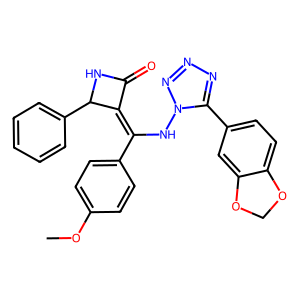} &
        \includegraphics[width=0.12\textwidth]{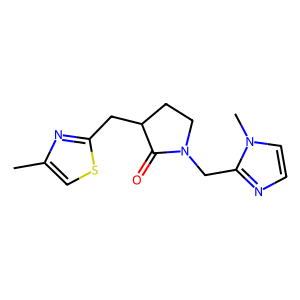} &
        \includegraphics[width=0.12\textwidth]{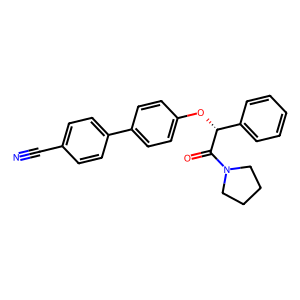} &
        \includegraphics[width=0.12\textwidth]{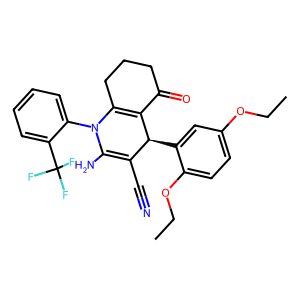}&
        \includegraphics[width=0.12\textwidth]{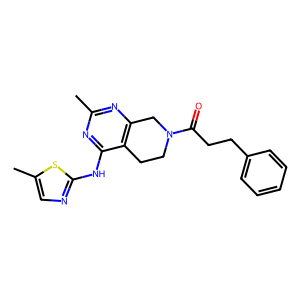} 
        \\
        \includegraphics[width=0.12\textwidth]{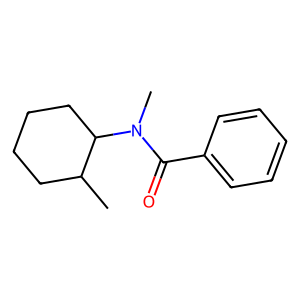} &
        \includegraphics[width=0.12\textwidth]{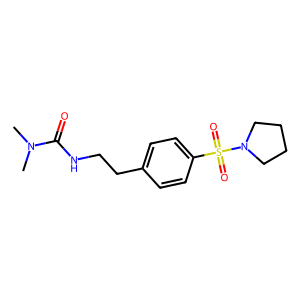} &
        \includegraphics[width=0.12\textwidth]{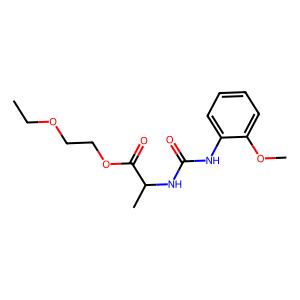} &
        \includegraphics[width=0.12\textwidth]{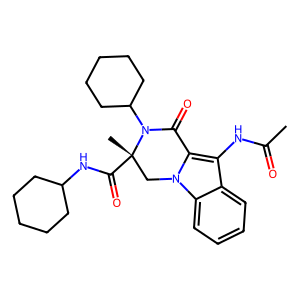} &
        \includegraphics[width=0.12\textwidth]{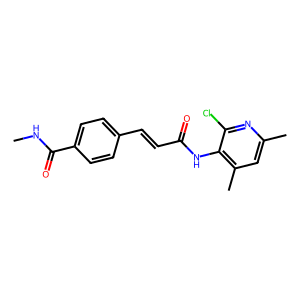}&
        \includegraphics[width=0.12\textwidth]{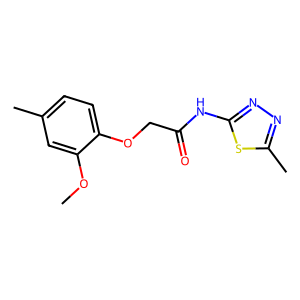} 
        \\
        \includegraphics[width=0.12\textwidth]{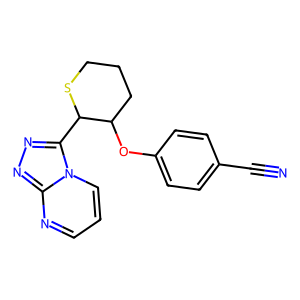} &
        \includegraphics[width=0.12\textwidth]{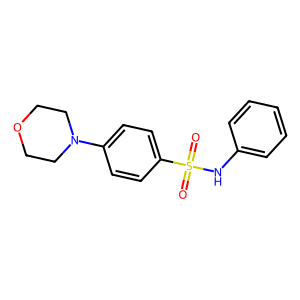} &
        \includegraphics[width=0.12\textwidth]{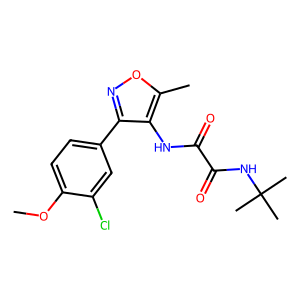} &
        \includegraphics[width=0.12\textwidth]{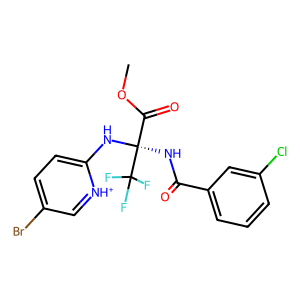} &
        \includegraphics[width=0.12\textwidth]{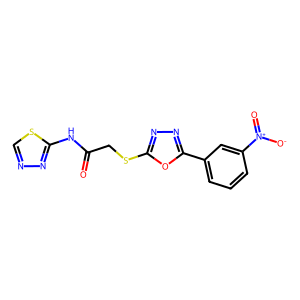}&
        \includegraphics[width=0.12\textwidth]{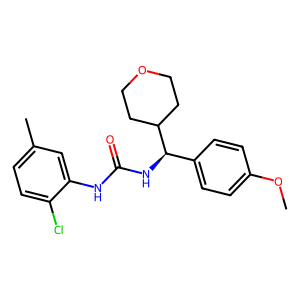} 
        \\
        \includegraphics[width=0.12\textwidth]{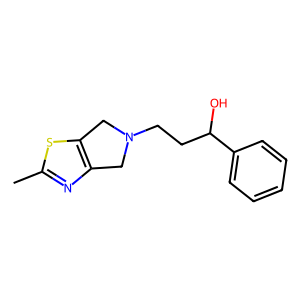} &
        \includegraphics[width=0.12\textwidth]{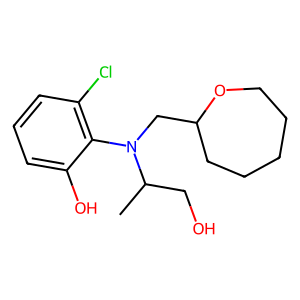} &
        \includegraphics[width=0.12\textwidth]{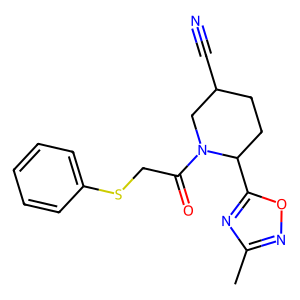} &
        \includegraphics[width=0.12\textwidth]{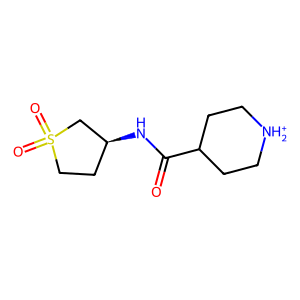} &
        \includegraphics[width=0.12\textwidth]{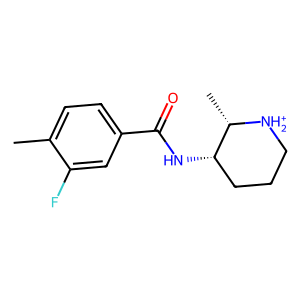}&
        \includegraphics[width=0.12\textwidth]{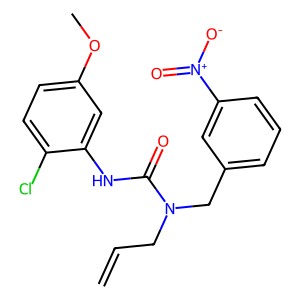} 
        \\
        \hline
        \end{tabular}
\end{figure}

\subsection{Generic Graphs}
\textcolor{white}{-}

\begin{figure}[H]
    \centering
    \caption{Planar}
    \begin{tabular}{|ccc|ccc|}
    \hline
      \multicolumn{3}{|c|}{Generated graphs}   &
      \multicolumn{3}{|c|}{Real graphs}
     \\
    \hline
           
        \includegraphics[width=0.15\textwidth]{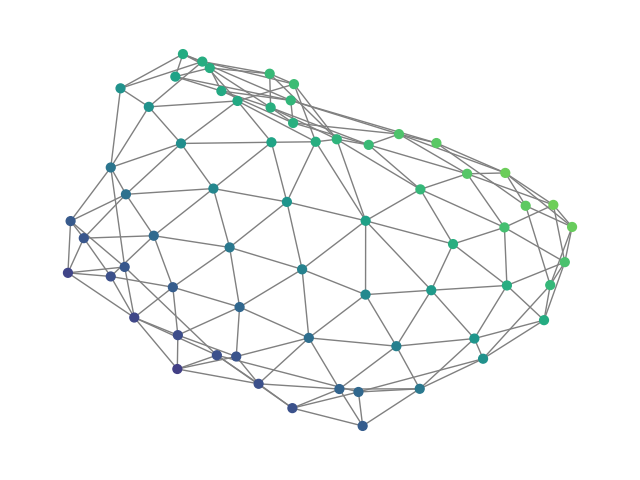} &
        \includegraphics[width=0.15\textwidth]{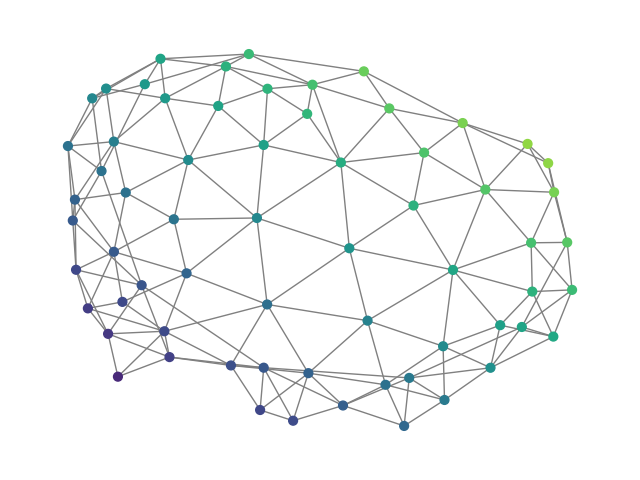} &
        \includegraphics[width=0.15\textwidth]{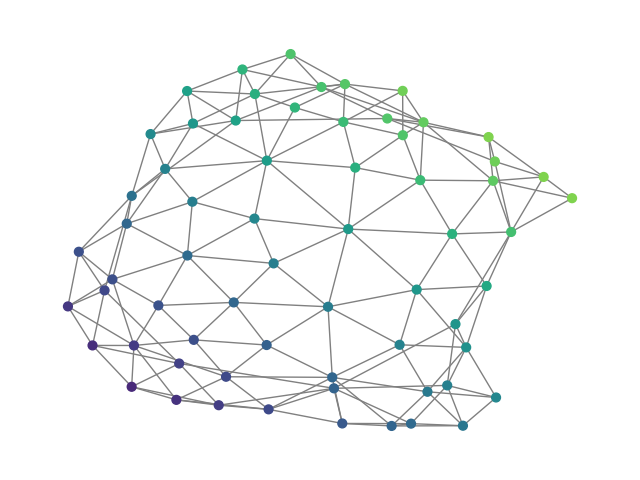} &
        \includegraphics[width=0.15\textwidth]{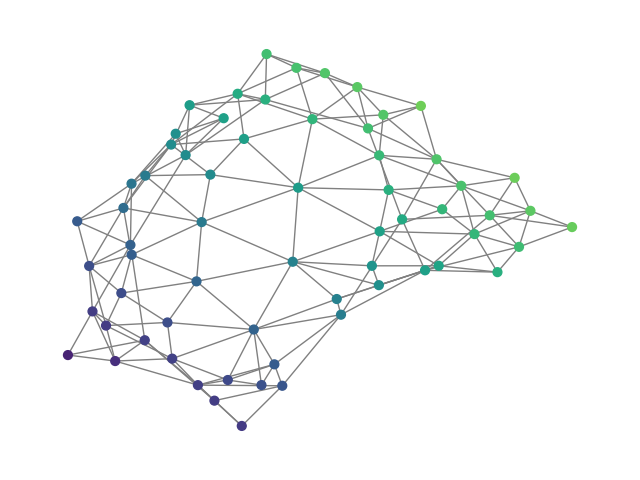} &
        \includegraphics[width=0.15\textwidth]{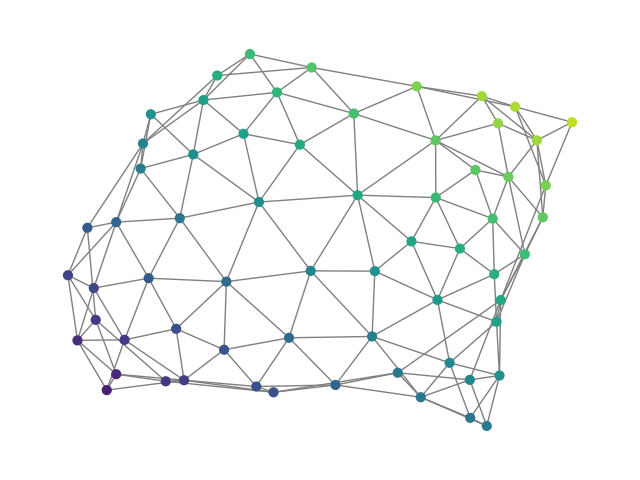}&
        \includegraphics[width=0.15\textwidth]{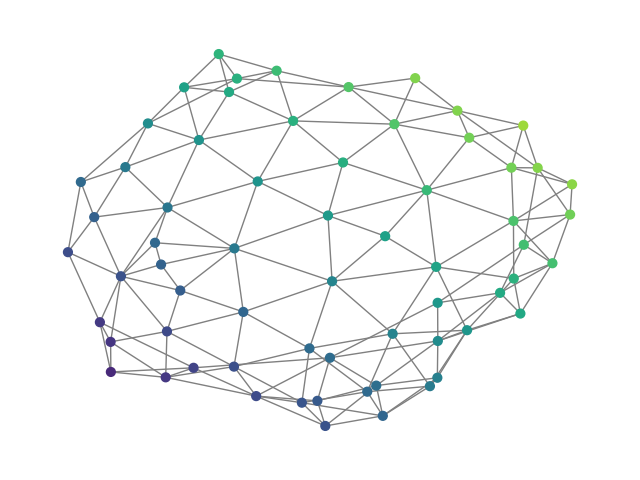} 
        \\
        \includegraphics[width=0.15\textwidth]{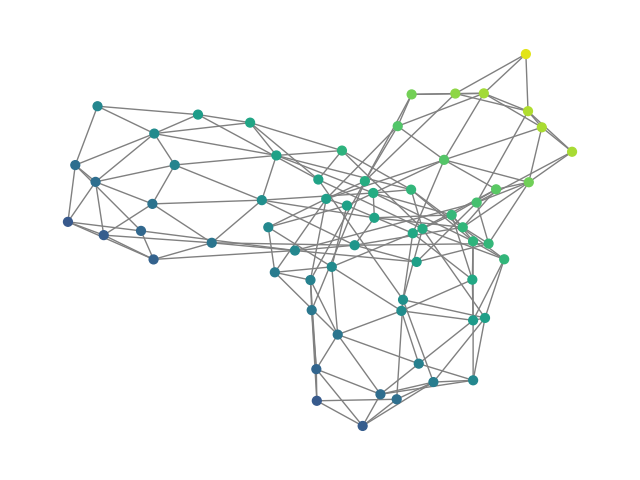} &
        \includegraphics[width=0.15\textwidth]{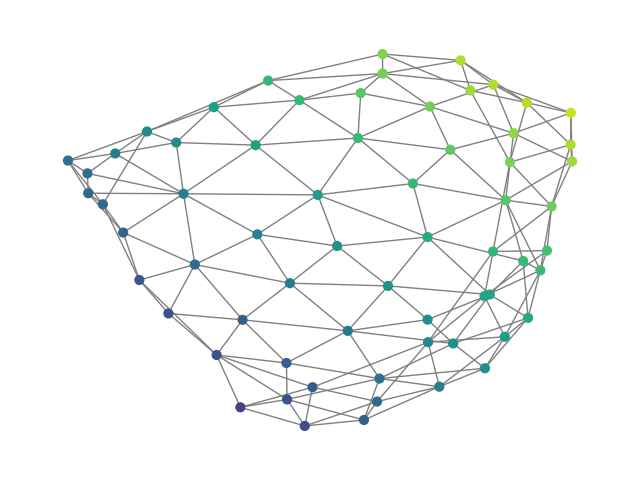} &
        \includegraphics[width=0.15\textwidth]{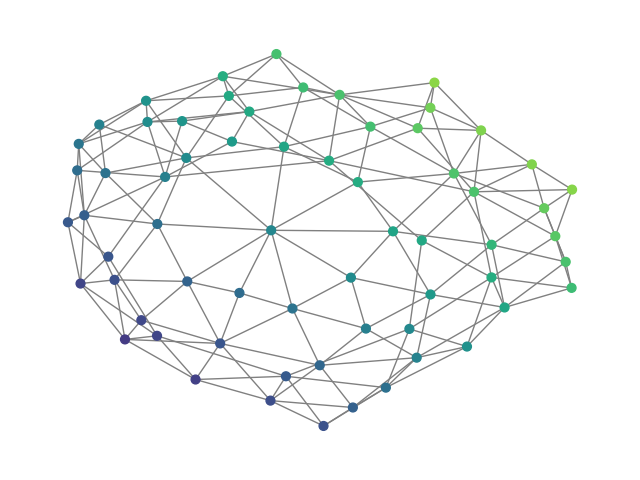} &
        \includegraphics[width=0.15\textwidth]{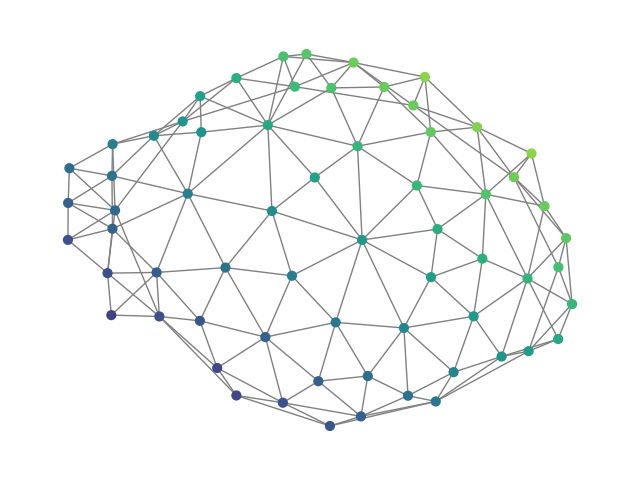} &
        \includegraphics[width=0.15\textwidth]{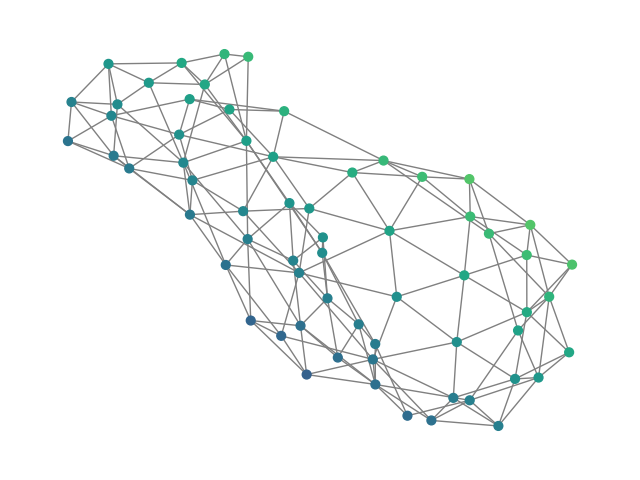}&
        \includegraphics[width=0.15\textwidth]{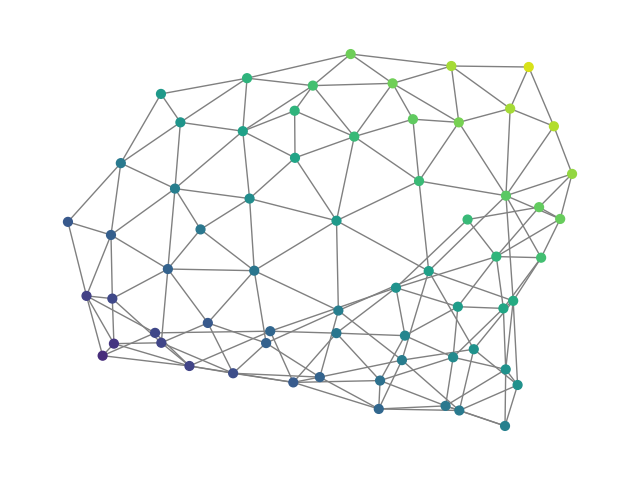} 
        \\
        \includegraphics[width=0.15\textwidth]{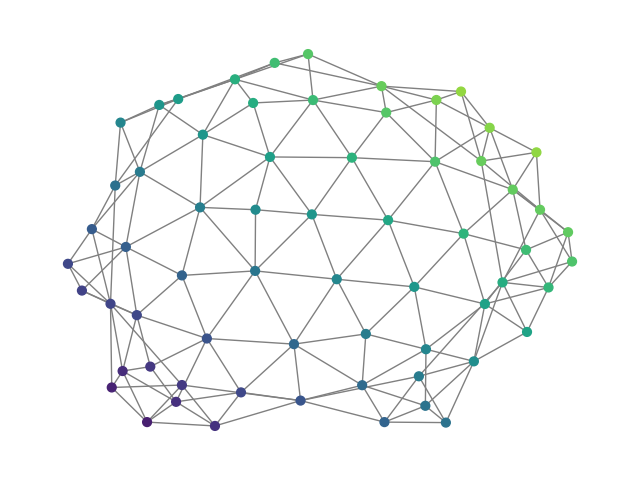} &
        \includegraphics[width=0.15\textwidth]{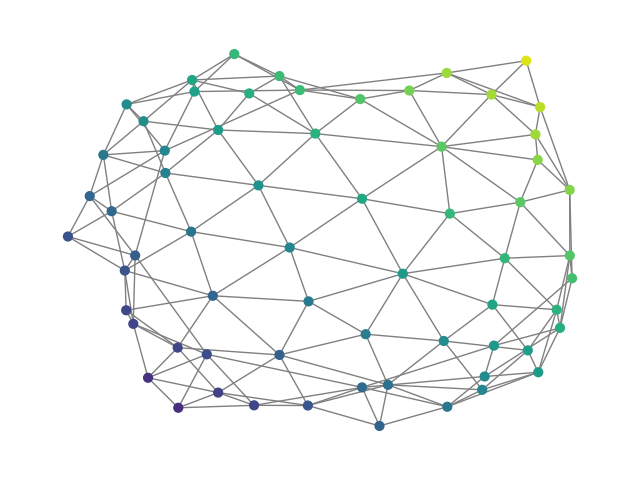} &
        \includegraphics[width=0.15\textwidth]{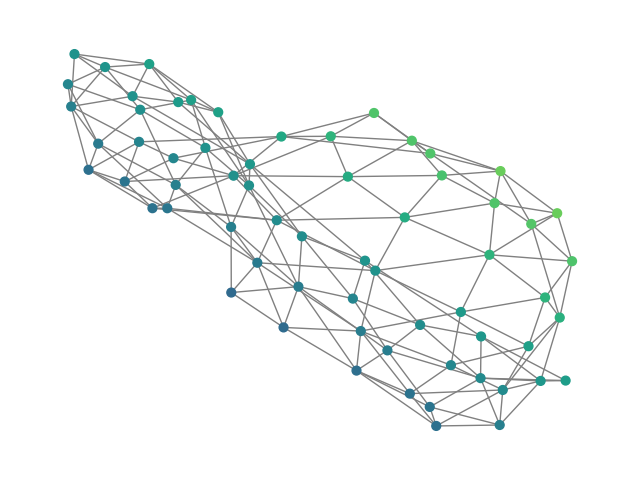} &
        \includegraphics[width=0.15\textwidth]{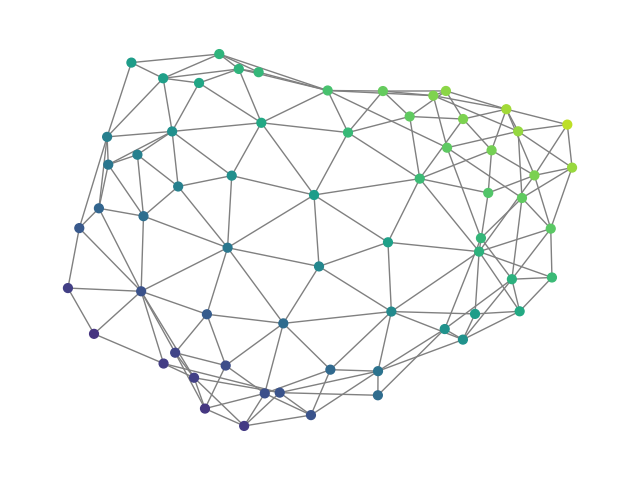} &
        \includegraphics[width=0.15\textwidth]{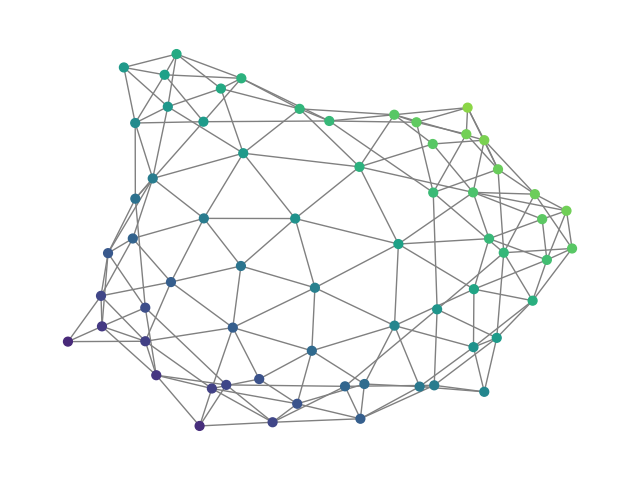}&
        \includegraphics[width=0.15\textwidth]{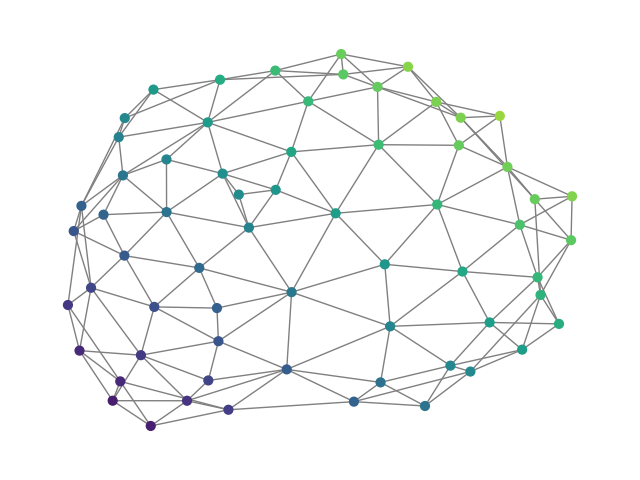} 
        \\
        \includegraphics[width=0.15\textwidth]{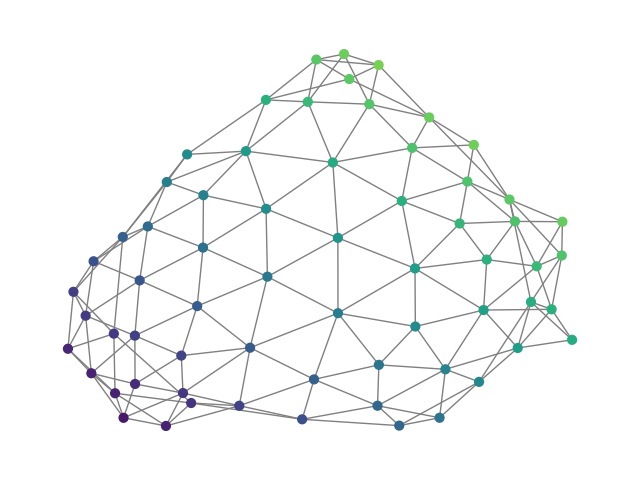} &
        \includegraphics[width=0.15\textwidth]{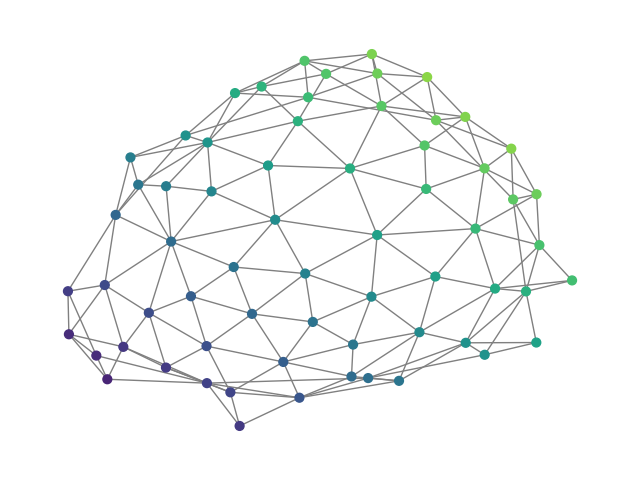} &
        \includegraphics[width=0.15\textwidth]{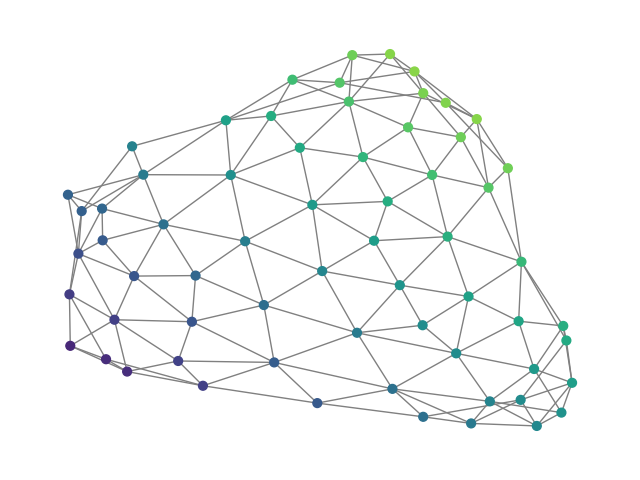} &
        \includegraphics[width=0.15\textwidth]{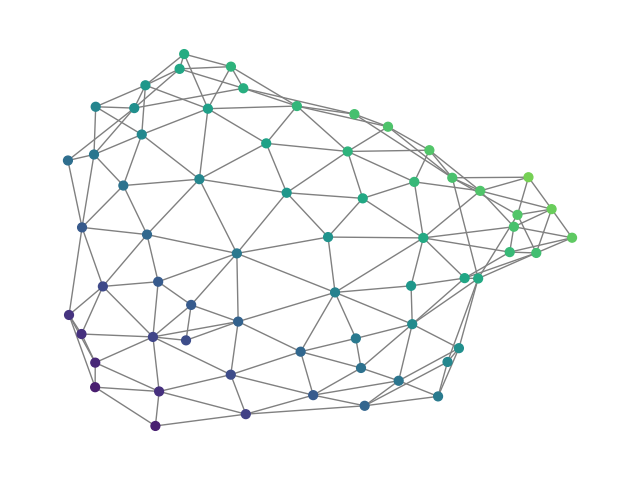} &
        \includegraphics[width=0.15\textwidth]{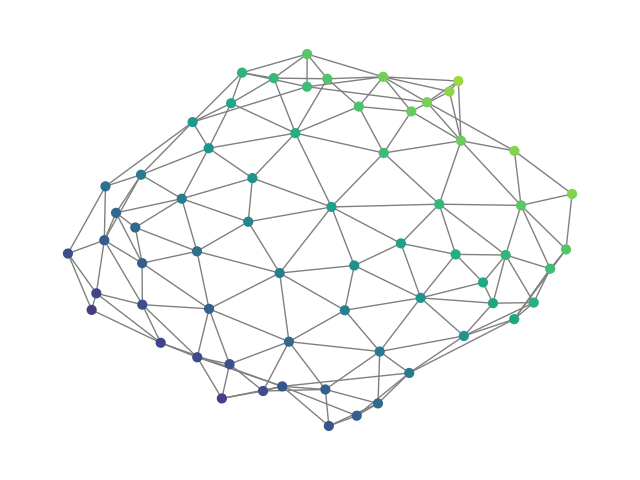}&
        \includegraphics[width=0.15\textwidth]{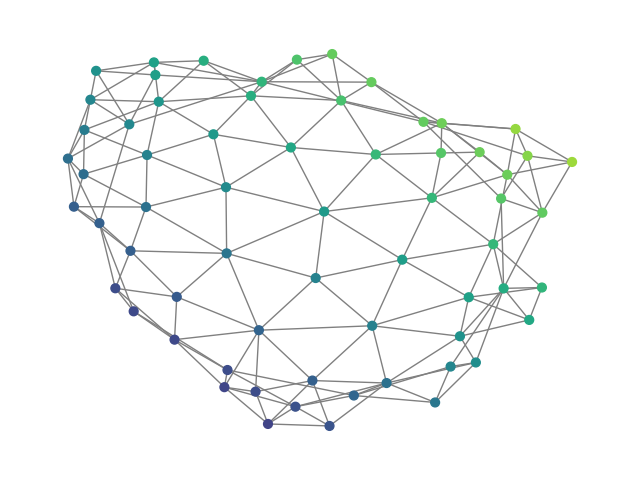} 
        \\
        \hline
        \end{tabular}
\end{figure}

\begin{figure}[H]
    \centering
    \caption{SBM}
    \begin{tabular}{|ccc|ccc|}
    \hline
      \multicolumn{3}{|c|}{Generated graphs}   &
      \multicolumn{3}{|c|}{Real graphs}
     \\
    \hline
           
        \includegraphics[width=0.15\textwidth]{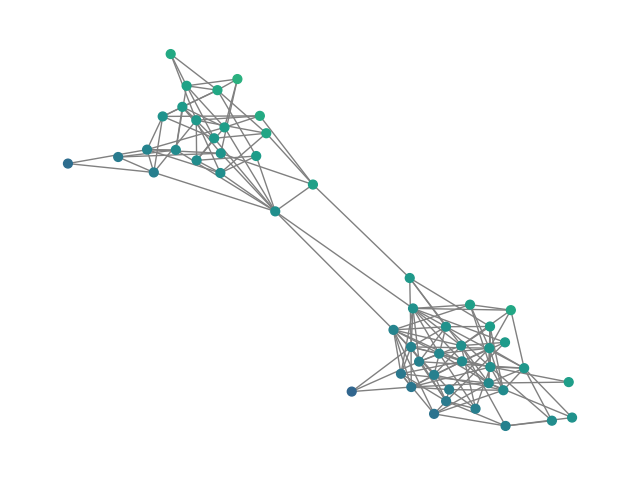} &
        \includegraphics[width=0.15\textwidth]{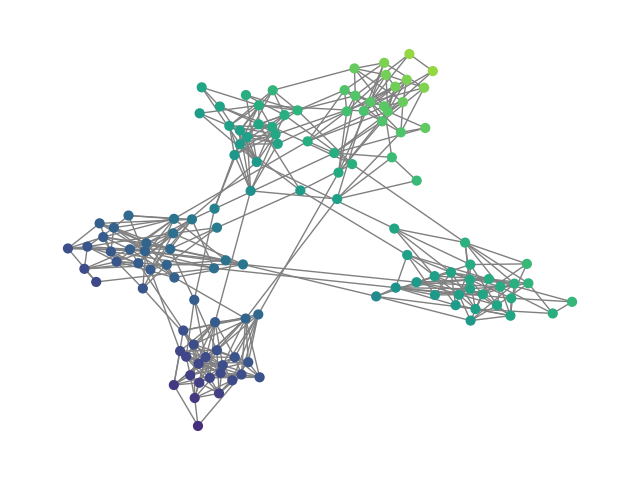} &
        \includegraphics[width=0.15\textwidth]{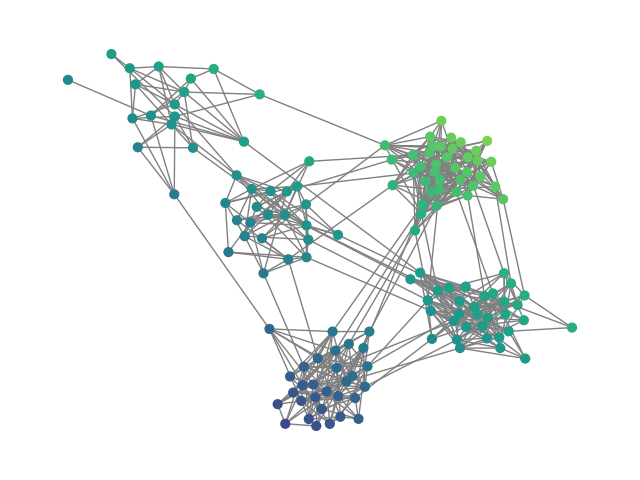} &
        \includegraphics[width=0.15\textwidth]{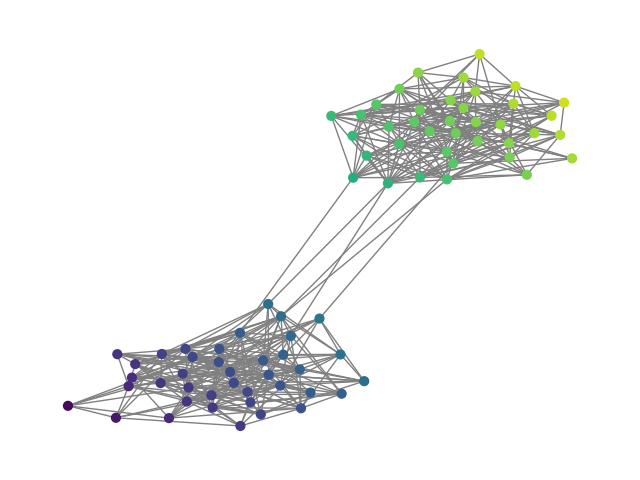} &
        \includegraphics[width=0.15\textwidth]{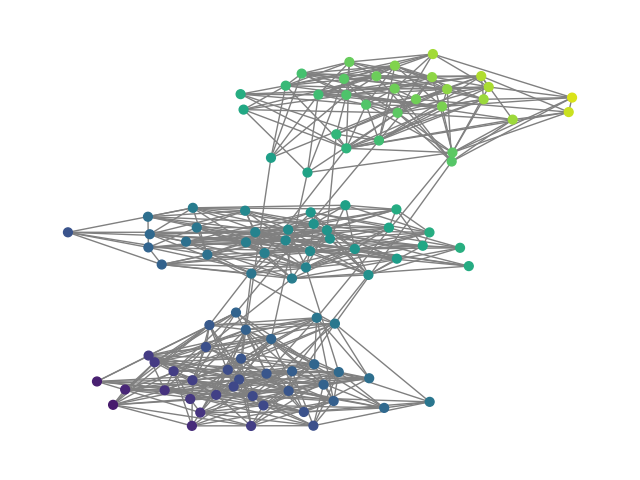}&
        \includegraphics[width=0.15\textwidth]{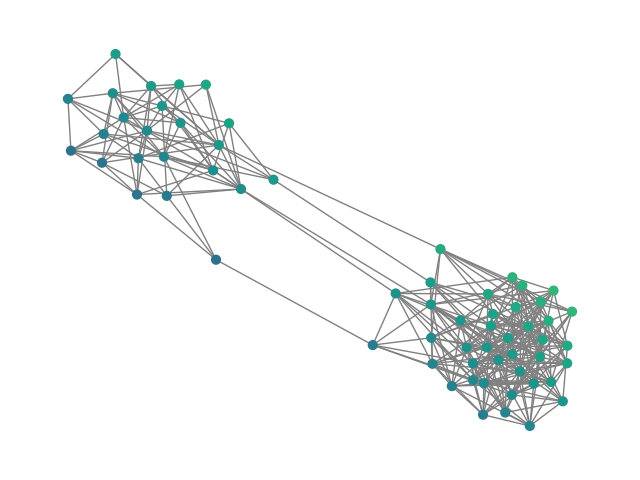} 
        \\
        \includegraphics[width=0.15\textwidth]{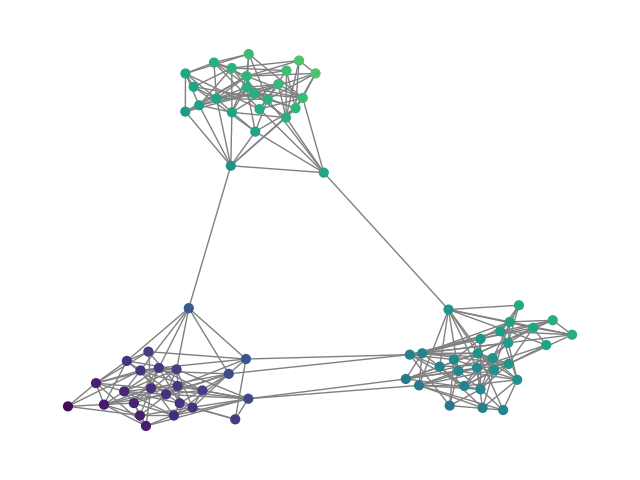} &
        \includegraphics[width=0.15\textwidth]{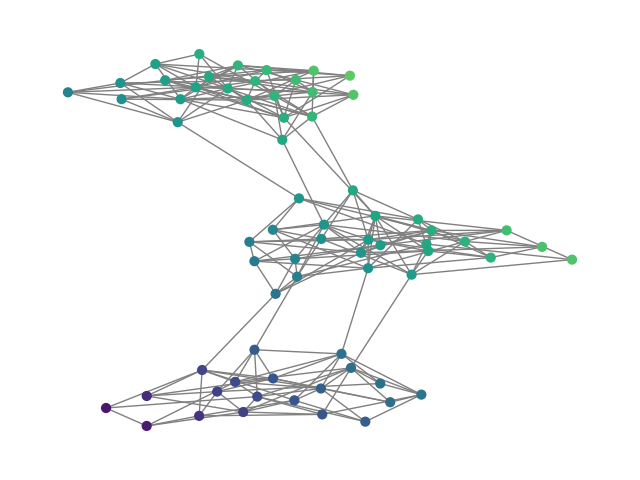} &
        \includegraphics[width=0.15\textwidth]{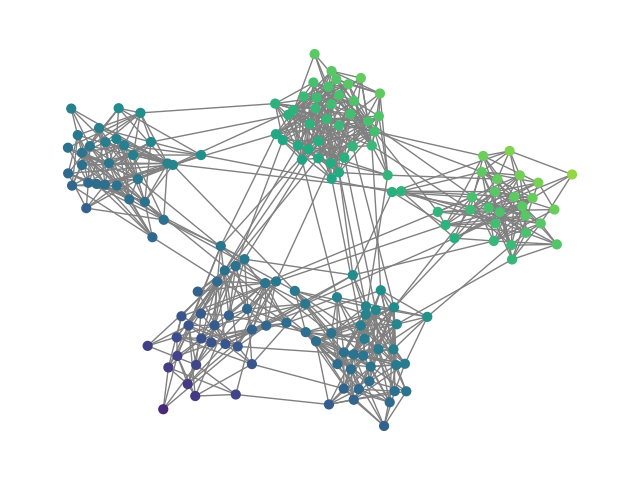} &
        \includegraphics[width=0.15\textwidth]{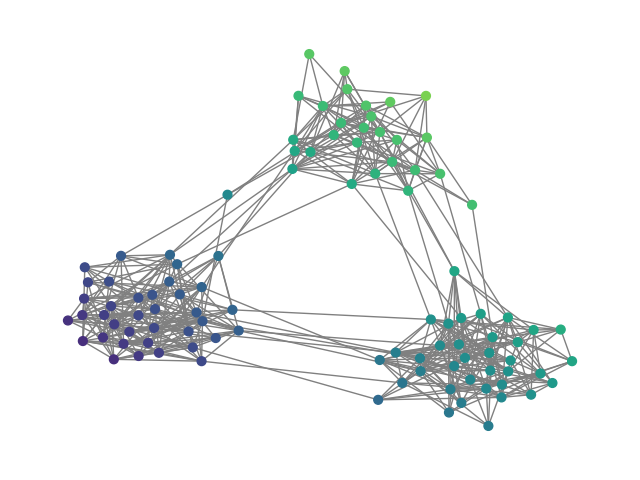} &
        \includegraphics[width=0.15\textwidth]{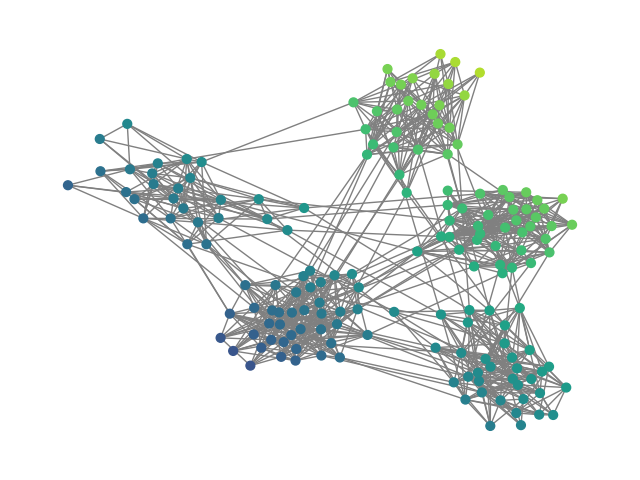}&
        \includegraphics[width=0.15\textwidth]{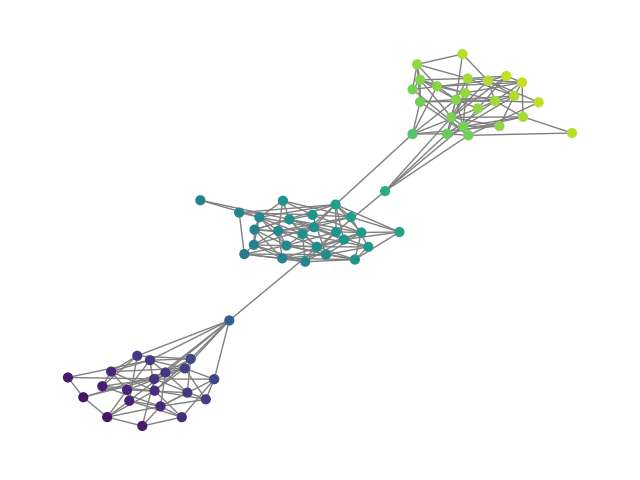} 
        \\
        \includegraphics[width=0.15\textwidth]{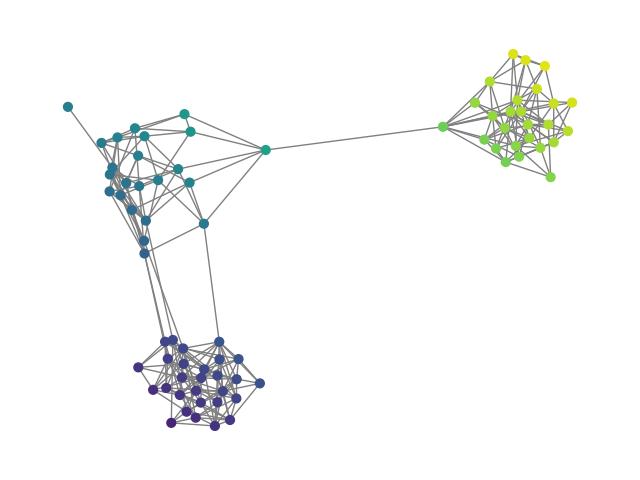} &
        \includegraphics[width=0.15\textwidth]{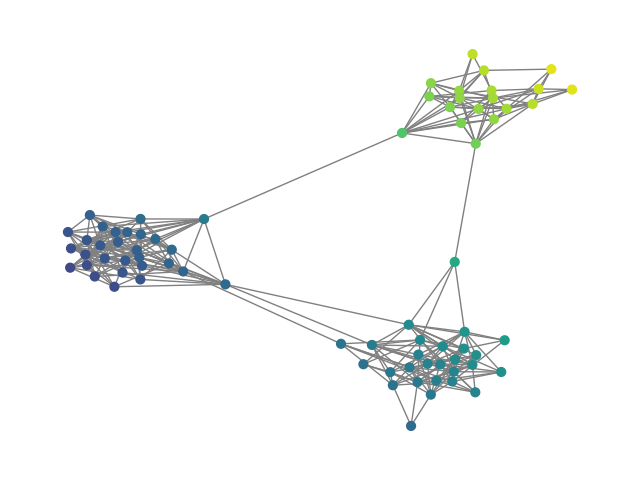} &
        \includegraphics[width=0.15\textwidth]{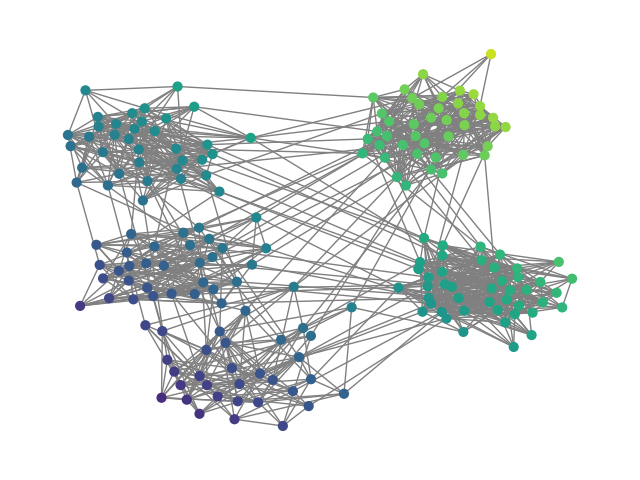} &
        \includegraphics[width=0.15\textwidth]{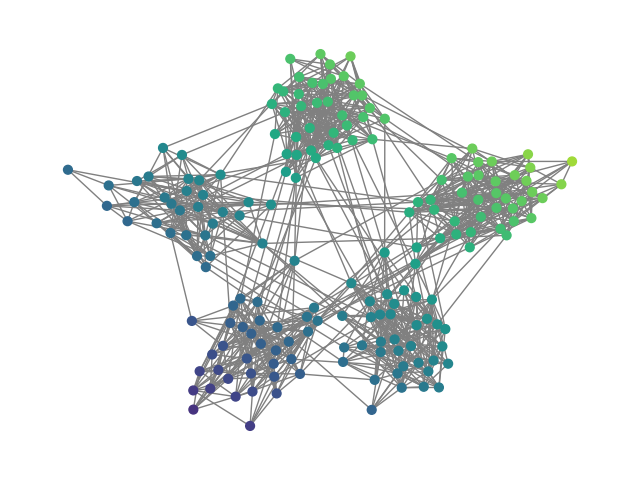} &
        \includegraphics[width=0.15\textwidth]{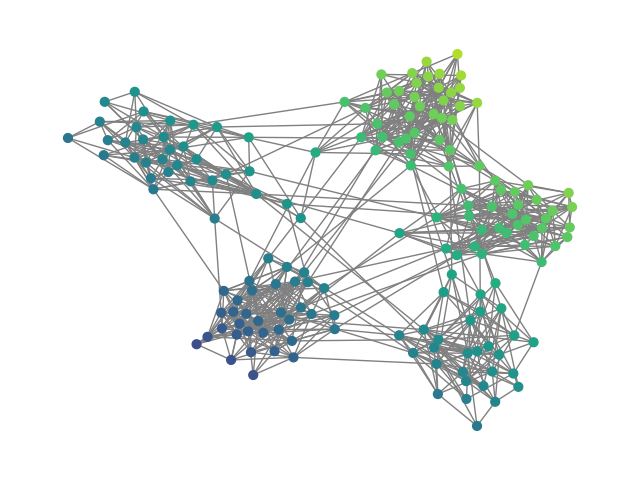}&
        \includegraphics[width=0.15\textwidth]{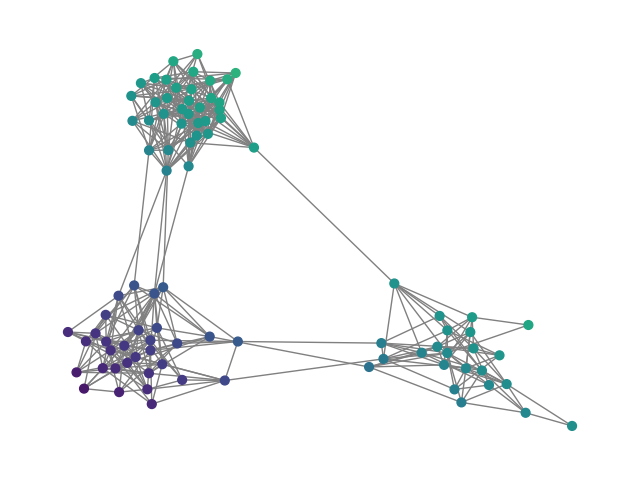} 
        \\
        \includegraphics[width=0.15\textwidth]{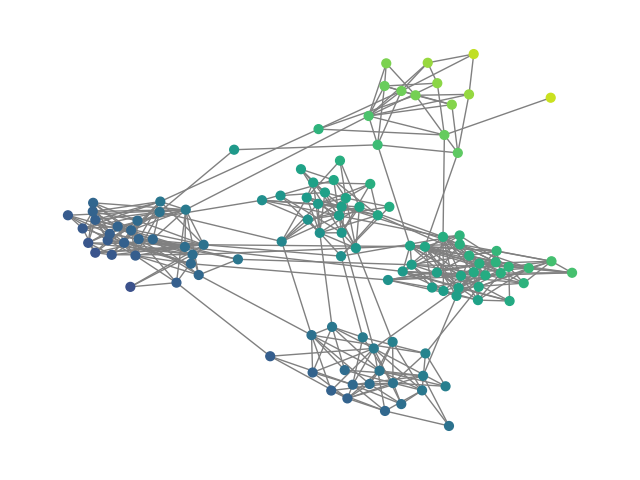} &
        \includegraphics[width=0.15\textwidth]{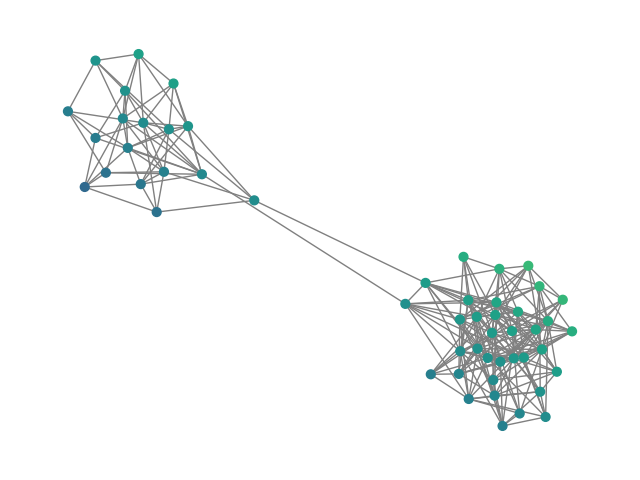} &
        \includegraphics[width=0.15\textwidth]{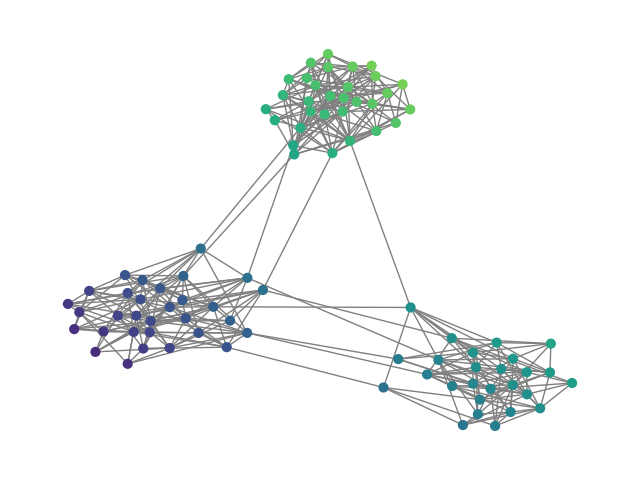} &
        \includegraphics[width=0.15\textwidth]{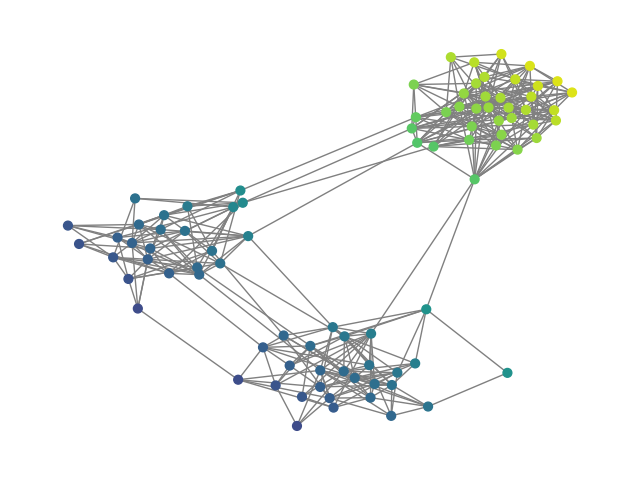} &
        \includegraphics[width=0.15\textwidth]{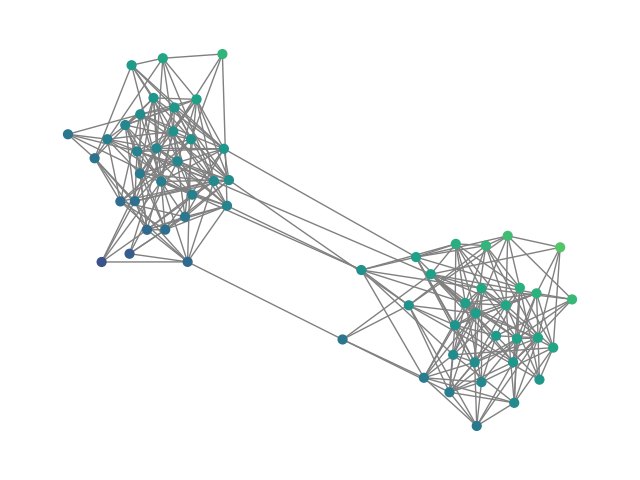}&
        \includegraphics[width=0.15\textwidth]{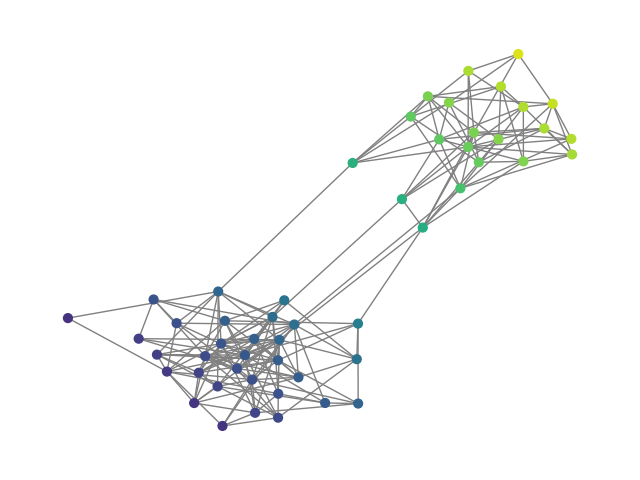} 
        \\
        \hline
        \end{tabular}
\end{figure}
All generated graphs comes from the Marginal Iterative Denoising model.

\end{document}